%% file: main.tex
\documentclass{ntupaper}

\PassOptionsToPackage{numbers, compress}{natbib}
\usepackage{tabularx}
\usepackage{longtable}

\usepackage[utf8]{inputenc}
\usepackage[T1]{fontenc}
\usepackage{hyperref}
\usepackage{url}
\usepackage{booktabs}
\usepackage{amsfonts}
\usepackage{nicefrac}
\usepackage{microtype}
\usepackage{xcolor}
\usepackage{booktabs}
\usepackage{multirow}
\usepackage{amsmath}
\usepackage{dsfont}
\usepackage{algorithm}
\usepackage{algpseudocode}
\usepackage{graphicx}
\usepackage{subcaption}
\usepackage{multicol}
\usepackage{wrapfig}
\usepackage{tikz}
\usepackage{xspace}
\usepackage[normalem]{ulem}
\usepackage{soul}
\usepackage{mathtools}
\usepackage{booktabs}
\usepackage{multirow}
\usepackage{amssymb}
\usepackage{threeparttable}
\usepackage{booktabs,xcolor,colortbl,array}

\usepackage{listings}
\lstdefinestyle{prog}{language=Python, basicstyle=\ttfamily\scriptsize, keywordstyle=\color{blue!60!black}\bfseries,
  commentstyle=\color{gray}\itshape, stringstyle=\color{red!50!black}, numbers=left, numberstyle=\tiny\color{gray},
  numbersep=6pt, frame=lines, framerule=0.4pt, breaklines=true, breakatwhitespace=true, showstringspaces=false,
  columns=fullflexible, keepspaces=true, tabsize=4, xleftmargin=1.5em, aboveskip=6pt, belowskip=6pt, literate={±}{{$\pm$}}1}
\input{content_v2/macros}

\runningtitle{}

\runningstatus{Under review}

\github{IDEAgent}
\correspondence{Soujanya Poria (\email{soujanya.poria@ntu.edu.sg})}
\date{\today}

\title{\IDEAgentLogo: Agentic Quality-Diversity Search for Research Idea Generation}

\input{content_v2/authors}
\input{content_v2/abstract}
\begin{document}
\maketitle
\input{content_v2/introduction}
\input{content_v2/related_works}
\input{content_v2/methodology}
\input{content_v2/data}
\input{content_v2/evaluation}
\input{content_v2/experiments}
\input{content_v2/discussion}
\input{content_v2/verifiable}

\input{content_v2/conclusion}
\input{content_v2/extra}

\bibliographystyle{plainnat}
\bibliography{references}

\appendix
\input{content_v2/appendix/limitations}
\input{content_v2/appendix/extra_results}

\input{content_v2/appendix/cost}
\input{content_v2/appendix/metrics}
\input{content_v2/appendix/dataset}

\input{content_v2/appendix/alpharesearch}

\end{document}

%% file: content_v2/macros.tex
\newcommand{\framework}{\textsf{IDEAgent}\xspace}
\newcommand{\yield}{$\mathrm{Yield}$\xspace}

\providecommand{\DIV}{D}
\providecommand{\SO}{S}
\providecommand{\NB}{\mathrm{NB}}
\providecommand{\CL}{C}

\newcommand{\geminipro}{\texttt{gemini-3.1-pro}\xspace}
\newcommand{\geminiflash}{\texttt{gemini-3.5-flash}\xspace}
\newcommand{\geminiflashold}{\texttt{gemini-2.5-flash}\xspace}
\newcommand{\gpt}{\texttt{gpt-5.6-sol}\xspace}
\newcommand{\opusold}{\texttt{claude-opus-4.7}\xspace}
\newcommand{\opus}{\texttt{claude-opus-4.8}\xspace}
\newcommand{\opusnew}{\texttt{claude-opus-5.5}\xspace}
\newcommand{\sonnet}{\texttt{claude-sonnet-5}\xspace}
\newcommand{\fable}{\texttt{claude-fable-5}\xspace}
\newcommand{\qwen}{\texttt{Qwen3.8-27B}\xspace}
\newcommand{\qwenmoe}{\texttt{Qwen3.6-35B-A3B}\xspace}
\newcommand{\gemma}{\texttt{Gemma-4-31B-it}\xspace}
\newcommand{\minimax}{\texttt{MiniMax-M2.7}\xspace}
\newcommand{\glm}{\texttt{GLM-4.5-Air}\xspace}
\newcommand{\alpharesearch}{\texttt{AlphaResearch}\xspace}
\newcommand{\alebench}{\texttt{ALE-Bench}\xspace}

\definecolor{ideacolor}{HTML}{0072B2}
\definecolor{agentcolor}{HTML}{D55E00}
\newcommand{\IDEAgentSharedA}{%
  \tikz[baseline=(box.base)]{%
    \node[inner sep=0pt,outer sep=0pt,font=\sffamily\bfseries,text opacity=0] (box) {A};
    \begin{scope}
      \clip (box.south west) rectangle (box.north);
      \node[inner sep=0pt,outer sep=0pt,font=\sffamily\bfseries,text=ideacolor] at (box.center) {A};
    \end{scope}
    \begin{scope}
      \clip (box.south) rectangle (box.north east);
      \node[inner sep=0pt,outer sep=0pt,font=\sffamily\bfseries,text=agentcolor] at (box.center) {A};
    \end{scope}
  }%
}
\DeclareRobustCommand{\IDEAgentLogo}{%
  \mbox{{\sffamily\bfseries\color{ideacolor}IDE}\IDEAgentSharedA{\sffamily\bfseries\color{agentcolor}gent}}%
}

%% file: content_v2/authors.tex
\author[1]{Varun Gumma}
\author[1]{Navonil Majumder}
\author[1]{Soumitra Sinhahajari}
\author[1]{Soujanya Poria}
\affiliation[1]{DeCLaRe Lab, Nanyang Technological University, Singapore}

%% file: content_v2/abstract.tex
\abstract{
Large Language Models (LLMs) have significantly advanced the automation of scientific discovery. However, existing systems share a core limitation: they optimize ideas independently for either \textit{Quality} or \textit{Diversity}. This often results in closely related ideas or large sets of trivial, unsound, or unclear concepts. We instead argue that research ideation should be framed as a conjunction of \textit{Quality} and \textit{Diversity} objectives, i.e., a \textit{Quality-Diversity} \textbf{(QD)} search. Accordingly, we introduce \framework, a \textit{multi-agent} framework that tracks idea evolution through \textit{lineages}. We jointly drive \textit{Quality} using multi-objective feedback for repair and refinement, while \textit{Diversity} is achieved through \textit{lightweight} sequential memory and explicit comparison against completed ideas, historical variants, and rejected proposals. To evaluate this joint objective, we develop \yield, a metric that computes the largest mutually diverse set of ideas satisfying predefined quality thresholds. Across 32 topics spanning 8 \texttt{CS} domains, \framework outperforms the best baseline by $3.89\times$ on \yield, while achieving \yield of at least $2$ on $8\times$ as many topics. We further analyze quality improvements, showing that repair and refinement improve logical rigor and clarity while preserving non-obviousness. Lastly, we extend \framework to two algorithm discovery tasks for verifiable quality judgment and find that it still consistently outperforms all baselines, and yields the largest set of high-quality and diverse programs that beat a trivial/standard baseline. To encourage future research on QD-search-based ideation, we open-source \framework at \url{https://github.com/declare-lab/IDEAgent}.
}

%% file: content_v2/introduction.tex
\section{Introduction}
\label{sec:intro}

\begin{figure*}[t]
    \centering
    \includegraphics[width=0.315\textwidth]{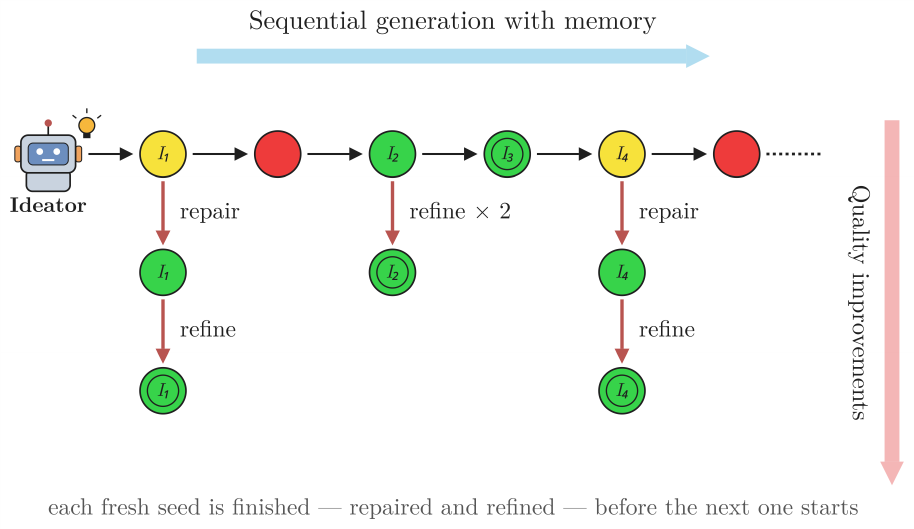}\hfill{\color{gray!50}\vrule width 0.6pt}\hfill
    \includegraphics[width=0.315\textwidth]{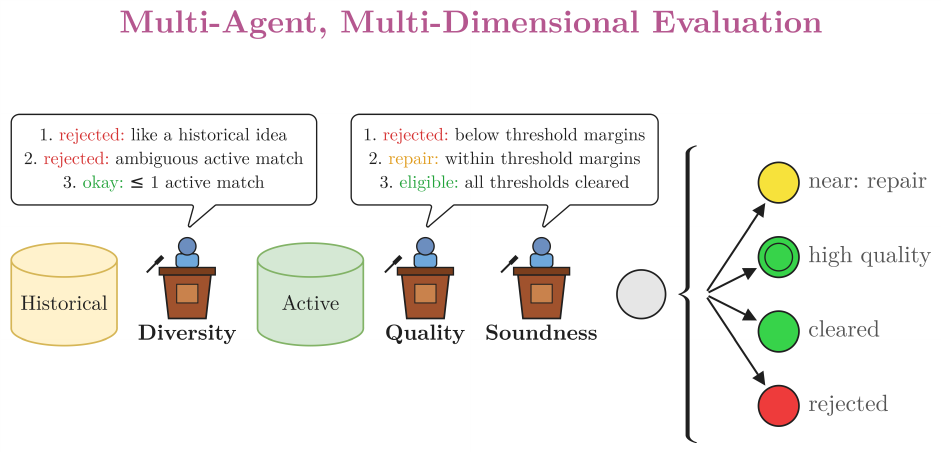}\hfill{\color{gray!50}\vrule width 0.6pt}\hfill
    \includegraphics[width=0.315\textwidth]{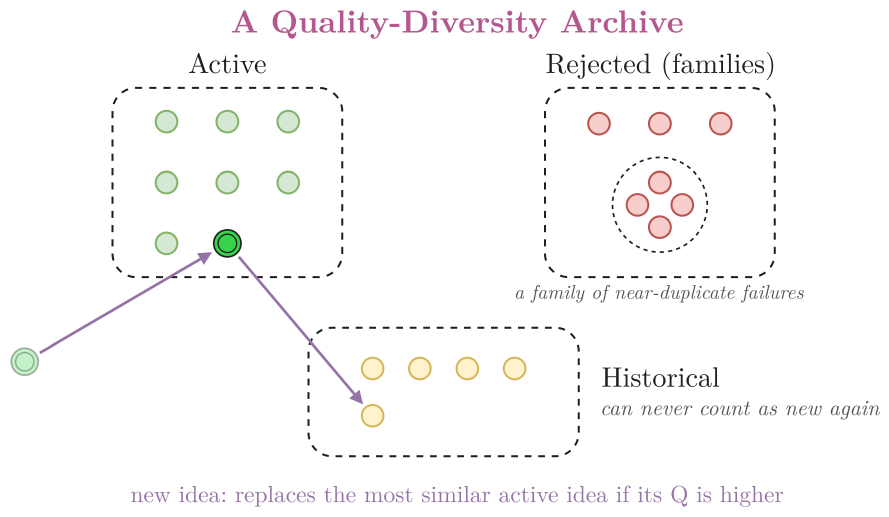}
    \caption{\framework at a glance. \textit{Left:} ideas are generated sequentially with memory, and each fresh seed is repaired and refined before the next one starts. \textit{Middle:} diversity, quality and soundness judges decide whether an idea is rejected, repaired or accepted. \textit{Right:} accepted ideas enter a quality-diversity archive of active, historical and rejected ideas.}
    \label{fig:title}
\end{figure*}

The advent of LLMs has garnered interest in automatic scientific discovery. Recent works have shown the abilities of LLMs in assisting with literature surveys, hypothesis or idea generation, experiment design, draft writing, and now even autonomously executing chunks of a comprehensive research or experiment pipeline \citep{lu2024aiscientistfullyautomated, coscientist, yamada2025aiscientistv2workshoplevelautomated}. Of the aforementioned, the most fundamental step is \textit{ideation}, where the core \textit{gap} is identified in the background, one or more \textit{research questions} \citep{sinhahajari2026limitsllmasjudgescientificnovelty} are formulated along with \textit{potential solutions}. In this work, we propose to view the ideation process as a \textit{Quality-Diversity Search} problem. The objective is to discover a \textit{set} of research ideas, where each proposal is \textit{non-obvious}, \textit{sound}, and \textit{clear} (\textbf{Quality}) while remaining mutually orthogonal (\textbf{Diversity}), allowing researchers to explore multiple independent research directions.

Despite recent progress, LLM-based ideation systems do not treat ideation as a rigorous QD search. Existing methods, from one-shot prompting to multi-agent systems, optimize each generation independently and evaluate ideas in isolation. Consequently, as more ideas are generated, successive generations frequently exhibit a \textit{conceptual collapse} by repeatedly revisiting similar regions of the design space or producing minor, paraphrased variations of earlier concepts. Hence, while these systems may generate strong individual ideas, they offer rapidly diminishing returns \citep{tang2026airesearchagentsnarrow} and are often riddled with inconsistencies while pushing for novelty. In real-world scientific workflows, an AI ideation system should generate not one proposal but a diverse set of \textit{independent}, \textit{sound}, \textit{clear}, and \textit{non-obvious} research directions.

Sustained human research naturally operates as a QD search. In each brainstorming session, researchers iterate over a concept---correcting logical flaws, tightening assumptions, and improving its overall \textit{Quality}. This evolution forms a \textit{lineage}, where higher-quality versions replace earlier drafts. Across sessions, diversity relies on recording three types of ideas: \textit{completed} concepts, permanently rejected \textit{dead ends}, and \textit{historical variants} superseded by later refinements. Filtering against these archives forces exploration into new and unexplored directions, maximizing \textit{Diversity}.

Motivated by this perspective, we introduce \framework, a multi-agent framework to navigate the QD search for research ideation. Our framework sequentially develops new ideas against compact memories of prior generations. Each idea is then evaluated by multiple specialized agents, while being compared against completed, historical, and rejected ideas for novelty, each with its dedicated archives. Accepted ideas may optionally be refined, while those that miss by a limited margin are put through repairs targeting exact flaws and inconsistencies. Finally, all comparisons operate over structured summaries for efficient semantic contrast.

Evaluating the QD search framework requires accounting for both objectives jointly. Independently, both \textit{Quality} and \textit{Diversity} metrics can be gamed. For example, a set of novel, but near-clone ideas represents a single contribution, while a diverse set of trivial, invalid, or unclear ideas holds no exploration or scientific potential. To address this loophole, we introduce \yield, a group-level metric that jointly measures quality in terms of non-obviousness, soundness, clarity, and diversity with pairwise distinctness. After scoring each requirement independently, \yield employs hard thresholds on the \textit{quality} axes before extracting the \textit{largest subset} in which all the ideas pairwise satisfy the diversity constraint. Physically, \yield is designed to measures the number of ideas a human researcher could \textit{potentially} pursue from the generated set.

We evaluate \framework on 32 research topics spanning 8 Computer Science domains, and show that it consistently produces a \textit{higher density of high-quality ideas within a limited ideation budget}. We validate our hypothesis that \framework achieves a higher average \yield and a greater number of topics with non-zero \yield than all considered baselines. We further analyze improvements across individual quality rubrics using both internal and external evaluators. Although the absolute scores differ between the two judges, they consistently agree on the direction of the improvements, particularly in soundness/rigor, clarity, and non-obviousness. Finally, we also extend \framework to verifiable algorithm discovery where the output performance acts as the surrogate for quality, allowing us to relax the intermediate subjective evaluators, yet retaining the best performance across all baselines.

In short, our contributions are as follows: 
\begin{enumerate}
    \item We formulate automated scientific ideation/discovery as a QD search and argue that systems must develop multiple high-quality ideas that are sufficiently diverse from one another to maximize the number of \textit{pursuable} research directions.
    \item We propose \framework, a multi-agent framework that maintains idea \textit{lineages} using dedicated quality evaluators. We further introduce separate archives for ideas at different life-cycle stages to enable efficient novelty comparison using compact core summaries instead of full textual ideas. Finally, we incorporate targeted repair and refinement routines driven by these evaluations.
    \item We introduce \yield, a joint metric to assess both the \textit{Quality} and \textit{Diversity} of a set of ideas. We complement this with extensive experiments across 32 research topics, demonstrating the effectiveness of \framework for the QD search problem by consistently generating a larger set of non-obvious, sound, clear, and mutually distinct ideas.
    \item Lastly, we also extend our framework to verifiable domains with two algorithm-discovery benchmarks for objective quality evaluation and find that it still outperforms all baselines, an generated the largest pool of diverse ideas at all quality cut-offs.
\end{enumerate}

%% file: content_v2/related_works.tex
\section{Related Work}
\label{sec:related_work}

\paragraph{Autonomous Frameworks for Scientific Discovery}
The role of LLMs in scientific discovery has rapidly shifted from drafting assistance to fully autonomous, end-to-end research pipelines \citep{lu2024aiscientistfullyautomated, coscientist, schmidgall-etal-2025-agent}. To improve hypothesis quality, early frameworks introduced structured multi-agent environments, demonstrating that collaborative debate among distinct agent personas yields more robust proposals than single-turn generation \citep{su-etal-2025-many, ueda-etal-2025-exploring}. To ground these proposals and prevent superficial suggestions, subsequent architectures incorporated retrieval loops that ground reasoning in the existing literature \citep{baek-etal-2025-researchagent, li-etal-2025-chain-ideas, zhao2025deepideationdesigningllm}. More recently, this paradigm has expanded from abstract ideation to active software execution that iteratively implements and optimizes code to solve competitive benchmarks \citep{yamada2025aiscientistv2workshoplevelautomated, toledo2026ai}.

\paragraph{Novelty Bottlenecks in AI-based Ideation}
Evaluating AI-based ideation pipelines across large corpora of hypotheses reveals a persistent failure of divergent thinking. LLM agents frequently exhibit conceptual narrowness, clustering outputs around the provided seed literature and recombining familiar technical variations rather than introducing genuinely new research directions \citep{tang2026airesearchagentsnarrow}. While \citet{hu-etal-2025-nova, 10.1145/3786335.3813161} propose methods for algorithmic divergence, achieving true open-ended exploration remains a critical challenge. Finally, empirical studies show that success on complex downstream engineering tasks is fundamentally bounded by the diversity of early-stage ideation \citep{audranreiss2025doesgoodairesearch}.

\paragraph{QD Search in Text Generation}
\citet{bradley2024qualitydiversity} apply an evolutionary algorithm that uses language models to generate mutations, evaluate candidates for elitism, and operate on a fixed archive grid to map the search space. In contrast, we maintain a more flexible evolutionary process without a fixed search grid or MAP-Elites, while improving idea quality through directed feedback.

%% file: content_v2/methodology.tex
\section{Methodology}
\label{sec:methodology}

\paragraph{Task Formulation as QD Search.} \framework formulates scientific ideation as a Quality-Diversity search, i.e., under a fixed discovery budget, the aim is to produce as many high-quality and mutually distinct \textit{ideas} as possible. Formally, given a background corpus $\mathcal{X}$ consisting of user-provided \textit{foundational} research papers targeting an area of interest, we define an \textit{idea} $I$ as a structured research proposal specifying its problem, causal diagnosis, intervention, mechanism, assumptions, expected effect, and evaluation strategy.

Standard QD methods organize the search space using predefined axes and cells \citep{bradley2024qualitydiversity, Lehman2024, 10.1145/2001576.2001606, mouret2015illuminatingsearchspacesmapping, 10.3389/frobt.2016.00040, NEURIPS2021_532923f1, Cully2015}. This is generally too restrictive to express free-form research ideas, which may combine multiple intervention points. We therefore leave idea generation unconstrained and judge diversity from the proposed problem and causal mechanism.

An idea generated in one shot may contain a promising mechanism, but could be inconsistent, obvious, or underspecified. We therefore repair and refine the seed idea while tracking its evolution as a lineage $\ell_I$. Close variants of the same mechanism inherit the same lineage and cannot be counted as separate discoveries. A seed and its lineage therefore represent one attempt to discover a mechanism and consume exactly one unit of the discovery budget $B$.

\paragraph{Agents and Search Memory.} The \textbf{Ideator} is the primary generator that reads a background corpus $\mathcal{X}$ to generate a seed idea or subsequently refine it based on targeted feedback. Since the initial proposal $I$ is loosely formed, a \textbf{Steno} \textit{summarizes} $I$ into a concise quintuple: $\sigma(I)=( p_I,m_I,v_I,a_I,e_I)$, where the fields denote the problem $p_I$, central mechanism $m_I$, novel value addition over the background literature $v_I$, key assumptions $a_I$, and expected measurable effect $e_I$. This decomposition provides a compact, interpretable representation for pairwise comparison between ideas.

Three evaluators assess each draft idea. The \textbf{Quality Evaluator} scores non-obviousness ($\NB_I$), mechanism clarity ($\CL_I$), and feasibility ($F_I$) on a 0--100 scale. \emph{Non-obviousness} measures whether the central value addition requires a genuine conceptual leap beyond routine extensions, direct combinations, or ablations. \emph{Mechanism clarity} measures whether the causal steps from intervention to expected effect are well described. \emph{Feasibility} measures whether a plausible implementation and evaluation path exists under realistic research resources. The \textbf{Soundness Panel} produces $M=5$ independently sampled judgments
$S_I^1,\ldots,S_I^M\in\{0,\ldots,9\}$ on the logical and mathematical consistency of the idea's arguments and assumptions. These multiple judgments mitigate rollout-specific biases or oversights \citep{ho2026soundnessbenchaiscientistreally}. Their mean forms a single representative score that is rescaled to 0--100 as $S_I=\frac{100}{9M} \sum_{j=1}^{M}S_I^j$. A judgment is \textit{severe} when the score is no more than one-third of the scale, i.e., $S_I^j\leq3$. At least one severe judgment marks the idea as \textit{disputed}, while three or more make it \textit{invalid}. Invalid ideas are discarded, whereas disputed ideas are retained for possible repairs. The \textbf{Diversity Judge} compares core signatures of the active, inactive, and rejected ideas against a freshly-generated idea. Based on the similarity with all previously qualified ideas, a Diversity score $D_I\in[0,100]$ is issued, along with a list of duplicates identifying the closest active and historical mechanisms for further comparison.

Finally, a \textbf{Critic} turns evaluator scores and failure evidence into focused feedback for repair or refinement of the current idea, rather than proposing a substitute direction.

The controller maintains four stores. The active archive ($\mathcal A$) contains currently accepted ideas. The repair queue ($\mathcal Q$) temporarily holds a promising near-miss during its immediate repair. The historical/inactive archive ($\mathcal H$) retains signatures of previously qualified ideas that have left $\mathcal A$, including replaced parents and ideas displaced by higher-quality ones. The rejected-pattern memory ($\mathcal R$) stores compact summaries of rejected ideas and recurring failure patterns, rather than full rejected ideas
to avoid context bloat from a large number of rejections. The subsequent seed prompts receive these compact representations
and the full parent idea is fed to the Ideator only for repair or refinement.

\paragraph{Search Procedure.} For each $b=1,\ldots,B$, the search completes a bounded exploration of one seed before generating the next: $I_b^{(0)}\longrightarrow[\textsc{Repair}] \longrightarrow[\textsc{Refine}]\longrightarrow I_{b+1}^{(0)}$, where the bracketed operations are conditional.

\subparagraph{Stage I -- Generate and assess.}
Each fresh-seed prompt asks the Ideator for a problem or mechanism distinct from all ideas represented in $\mathcal A$, $\mathcal H$, $\mathcal Q$, and $\mathcal R$. If the preceding lineage ends as a historical duplicate or matches with a rejected pattern/logic, the prompt is augmented with an explicit instruction to pivot to abandon that failing mechanism or problematic area in its next generation. Once a draft is produced, the Steno and evaluators produce its structured quintuple summary, quality scores, soundness panel, and archive-relative diversity judgment. Based on the evaluator assessments, we define the core qualification gate as: $G(I) = \mathds{1}\big[ (N_I \geq \tau_N \wedge S_I \geq \tau_S \wedge C_I \geq \tau_C) \wedge \neg\operatorname{invalid}(I) \wedge \neg\operatorname{disputed}(I)\big]$. 
The primary search uses $\tau_N=\tau_S=\tau_C=60$ and a diversity floor $\tau_D=60$. Feasibility is measured, but excluded from the gate, as an ambitious and sound mechanism may be valuable, despite its immediate viability.

\subparagraph{Stage II -- Route and repair.}
The Diversity Judge first determines the candidate's relation to $\mathcal{A}$ and $\mathcal{H}$, after which the controller applies four cases: \textbf{(1)} An idea is a \textit{purely historical match} if it matches only historical mechanisms, but not the active refined successors in their lineage. Such candidates are sub-optimal and rejected. Candidates matching multiple active ideas are also rejected because they cannot be assigned unambiguously to a single lineage. \textbf{(2)} A candidate matching exactly one active idea is treated as another version of that lineage rather than a new discovery. If it passes $G$, it may replace the matched idea based on the aggregated quality score; otherwise, it is rejected. \textbf{(3)} A candidate with no duplicate matches, but $D_I<\tau_D$ indicates a combination or reuse of multiple pre-qualified ideas without exactly matching one. Such candidates are rejected for insufficient diversity. Repair is ineffective here because it is instructed to preserve the central mechanism. \textbf{(4)} Finally, a candidate with $D_I\geq\tau_D$ and no duplicate matches is admitted to the active set if it passes $G$. Otherwise, if it falls within $\delta=20$ points of the non-obviousness, soundness, and clarity thresholds, and is not \textit{invalid} under the Soundness Panel evaluation, it is eligible for \textit{one} repair. A \textit{disputed} candidate may bypass the numerical soundness margin but must satisfy the same diversity and non-obviousness/clarity proximity requirements.

For repair, the Critic receives the full idea, its failed rubrics, quality evidence, soundness objections, and a list of competing or similar mechanisms. The Ideator then returns a corrected child within the same lineage. The child is re-evaluated and must satisfy the same qualification and diversity requirements. If the repair fails, the original seed is retained, and its failure is summarized in $\mathcal R$. 

Candidates that qualify compete in $\mathcal{A}$ according to $Q_b(I)=0.7N_I+0.2S_I+0.1C_I$. This weighting emphasizes the innate non-triviality of an idea, which is difficult to achieve through refinement or repair. A new mechanism enters $\mathcal A$ when capacity is available. If the archive is full, it replaces the weakest active idea only if its $Q_b$ is higher, with ties broken by Feasibility ($F_I$). When an accepted idea leaves $\mathcal A$, its summary is stored in $\mathcal H$, not $\mathcal R$, since it remains a valid, high-quality idea.

\subparagraph{Stage III -- Refine accepted ideas.}
An accepted idea is eligible for refinement while it remains below a soft target: $N_I<90$, $S_I<80$, or $C_I<80$. These targets serve only as refinement conditions and are not additional acceptance gates. The Critic addresses the relevant deficit, and the Ideator produces a child within the same lineage. During diversity assessment, its own parent is naturally excluded to avoid trivial lineage comparisons.

The child replaces its parent only if it passes $G$, clears
$\tau_D$ relative to foreign mechanisms, and improves $Q_b$. Further, it must also pass a blinded parent--child non-obviousness comparison performed twice with their positions exchanged to account for position bias. If both comparisons prefer the same candidate, the winner is retained; otherwise, position disagreement is treated as a tie. This safeguard prevents improvements in clarity or soundness from trivializing the central mechanism. A non-improving child is discarded and does not overwrite its accepted parent, while a replaced parent is saved in $\mathcal{H}$. Feasibility remains a soft check throughout the process and is used only to break ties when $Q_b$ matches. The Ideator provides a plausible implementation and evaluation path, but feasibility neither triggers repair or refinement nor is included in the Critic's directed feedback.

The primary configuration uses $B=10$ fresh seeds and active capacity $C_{\mathcal{A}}=10$. Repair and accepted refinement share an allowance of at most $K_{\mathrm{aux}}=2$ auxiliary drafts per seed, with at most one repair and two accepted refinements. A seed repaired once therefore has room for at most one subsequent refinement, whereas an immediately accepted seed may receive two. The search ends after all $B$ fresh seeds and the final auxiliary chain completes. The resulting $\mathcal A$ contains the final set of ideas generated by \framework, which are used for downstream evaluation.

%% file: content_v2/data.tex
\section{Background Corpus Collection}
\label{sec:data}

We follow the data collection procedure of \citet{sinhahajari2026limitsllmasjudgescientificnovelty} to extract background papers from multiple arXiv \texttt{CS} domains and build the background corpus $\mathcal{X}$. Our objective is to generate a diverse set of non-trivial, sound research ideas that could \textit{potentially} emerge from a shared set of background papers serving as a seed knowledge bank. To this end, we begin with published papers from each domain and identify the broader set of papers that \textit{might} have influenced them. For each domain, papers published within a predefined time frame are collected and filtered by publication status and ranking. We then independently extract their \textit{influential citations} (ICs) \citep{sinhahajari2026limitsllmasjudgescientificnovelty} and group papers according to IC overlap. Specifically, two papers belong to the same group if the Intersection-over-Union (IoU) of their ICs exceeds $0.5$. The union of all ICs in a group forms its set of \textit{background papers}. Next, we retain only groups of \textit{background papers} supporting at least \textit{three} published papers, ensuring that each knowledge bank has demonstrated the ability to support multiple high-quality research ideas. We define each such collection of background papers as a \textit{topic}, representing a uniquely identifiable sub-domain or task. Topics containing between 5 and 8 background papers are retained and assigned descriptive tags using an LLM. Finally, we set a budget of $\Gamma=6$ and apply greedy maximum diversity selection (\Cref{alg:greedy_diversity}) to obtain the final collection of topics is provided in \Cref{tab:domains_topics}.

%% file: content_v2/evaluation.tex
\section{Evaluation}
\label{sec:eval}

In line with our QD formulation of research ideation, we evaluate each method on the returned \emph{set} of $N$ ideas per topic rather than in isolation. A valid set must satisfy two complementary requirements: ideas must individually be high-quality while remaining mutually diverse. To capture these, our evaluation protocol uses an LLM judge to provide independent per-idea \textit{Quality} scores and pairwise \textit{Diversity} metrics, which are then combined as follows\footnote{The exact scale and descriptions are provided in Appendix \ref{app:metrics}}:

\paragraph{\textbf{Quality Axis}} We evaluate each idea along three primary$^{\dagger}$ and two secondary quality axes:
\begin{enumerate}
    \item \textbf{Non-obviousness$^{\dagger}$} ($\NB{}_I$) measures the originality of the core contribution. The judge evaluates the idea according to \textit{Problem-Mechanism pairing}, \textit{Causal Diagnosis}, and \textit{Intervention}, which assess whether the idea would \textit{surprise} a knowledgeable human researcher while ignoring presentation differences.
    
    \item \textbf{Soundness$^{\dagger}$} ($\SO{}_I$) measures whether the proposed mechanism works as claimed and whether the logical flow contains internal contradictions or relies on non-rigorous or untested assumptions. A safe, unoriginal idea can still score low if its reasoning contains gaps, while a novel idea can score high if its logic is internally consistent.
    
    \item \textbf{Mechanism Clarity} ($\CL{}_I$) measures how concretely and unambiguously the mechanism is specified, and whether someone else could implement it from the description without inventing missing pieces. This is evaluated independently of whether the mechanism is correct (soundness) or practical to build (feasibility).
    
    \item \textbf{Feasibility} measures how practical it would be to build and test the idea under realistic constraints including data, compute, existing tools, and time. It ignores whether the idea is correct or important, as an idea can be highly feasible yet unsound, or vice versa.
    
    \item \textbf{Significance} measures the potential impact if the idea worked exactly as claimed, both on the target problem and the broader field. The judge assumes the mechanism works as claimed and ignores its feasibility or maturity.
\end{enumerate}

\paragraph{\textbf{Diversity Axis}} Given a set of $N$ ideas, \textit{Diversity} is evaluated pairwise. Each idea is independently decomposed into a semantic signature along eight axes: \textit{failure mode, causal diagnosis, intervention, signal source, intervention locus, objective, evaluation regime,} and \textit{assumption class}. Each of the $\binom{N}{2}$ idea pairs is then compared along all eight axes in a single judgment, producing a same/variant/related/distinct relation for each axis. A second call takes these axis-level relations as fixed evidence and produces a single score $\DIV{}_{ij} \in \{0,\dots,9\}$ by weighting disagreement on the \textit{core} axes---\textit{failure mode, causal diagnosis,} and \textit{intervention}---more heavily than the remaining supporting axes. For each topic $T$, we report the average pairwise diversity as $D(T)=\binom{N}{2}^{-1}\sum_i\sum_{i<j}\DIV{}_{ij}$

\paragraph{Yield} Our primary indicator is \yield$(\NB \geq k, \SO \geq l, \CL \geq m, \DIV\geq\tau)$ (or $Y_{k,l,m,\tau}$), which counts the largest number of \textit{sound, clear, non-trivial}, and \textit{mutually diverse} research ideas in a generated set. We first retain ideas satisfying the thresholds on \textit{non-obviousness} ($k$), \textit{soundness} ($l$), and \textit{clarity} ($m$). We then extract the \textit{largest} subset (maximum clique via branch-and-bound, \Cref{alg:yield}) in which every idea pair satisfies $\DIV_{ij}\ge\tau$. Consequently, a set of $N$ highly novel and sound, yet near-identical ideas correctly yields a score of $1$ rather than $N$, penalizing methods that exploit quality by repeatedly paraphrasing a small number of successful concepts.

\paragraph{Successful Topics}
A topic $T$ is successful at yield level $\Phi$ for a method if it generates at least $\Phi$ ideas satisfying the \yield criteria. We therefore report the proportion of successful topics as $|\mathcal{T}|^{-1} \sum_{T\in \mathcal{T}}~\mathds{1}[Y^{T}_{k,l,m,\tau}\ge \Phi]$. An ideal idea generator achieves a ratio of $1$.

%% file: content_v2/experiments.tex
\section{Experimental Setup}
\label{sec:exps}

All experiments use proprietary large-scale models, with the latest \gpt{} serving as the Ideator. The Critic, Soundness Panel, and Quality Judge are built around \geminipro{}, while the Diversity Judge and Steno use \geminiflash{} and \geminiflashold{}, respectively. During evaluation, we use two external large-scale models, \sonnet{} and \opusold{}, for all judgments. During analysis, we average their scores and treat the result as a \textit{jury verdict} to account for judge bias and sampling stochasticity. Finally, we use the same idea budget ($B=10$) for all baselines and report aggregated scores across all 32 topics.

\paragraph{Stateless.}
The simplest generation strategy, in which the Ideator produces $B$ ideas independently without knowledge of one another.

\paragraph{One-Shot ($\mathcal{OS}$).}
The Ideator generates all $B$ ideas in a single output. With a thinking-enabled Ideator, the model can use its reasoning space as shared memory for developing and diversifying ideas. Furthermore, each idea naturally conditions on all preceding ideas through causal generation, analogous to a rudimentary form of \textit{stateful} generation.

\paragraph{Sequential-Memory ($\mathcal{SM}$).}
Building on this notion of statefulness, each of the $B$ ideas is generated independently while cumulatively carrying forward the core summaries of previous ideas, allowing the model to retain a lightweight memory of prior generations for diversification. This setup is sequentially similar to \framework, but omits the quality-improvement subroutines (repair and refinement).

\paragraph{NOVA-inspired $\mathcal{SM}$.}
Building on Sequential-Memory, we augment it with NOVA's \citep{hu-etal-2025-nova} iterative \textit{seed-pool}-based germination and replacement strategy for diversification. We discard NOVA's online retrieval module, since our background papers constitute a closed knowledge base. Specifically, we run Sequential-Memory for $3$ rounds, each generating $B$ ideas. At the end of every round, an external judge reflects on the generated ideas and selects the $3$ most promising as seed concepts, replacing those from the previous round. The next round germinates $B$ new ideas from these seed directions together with the accumulated global memory. Finally, the Critic selects the best $B$ ideas from the complete pool of $3B$ generations, emphasizing distinctness, to form the final evaluation set. As before, each idea is generated in a single attempt without subsequent repair or refinement.

%% file: content_v2/discussion.tex
\section{Results and Analysis}
\label{sec:analysis}
\input{content_v2/tables/headline_table_main}

All the scores and numbers we present here are aggregated across all 32 topics.

\paragraph{\framework has a Higher Yield}

\begin{figure}[t]
    \centering
    \begin{subfigure}[b]{0.395\linewidth}
        \centering
        \includegraphics[width=\linewidth, trim=0 0 0 18pt, clip]{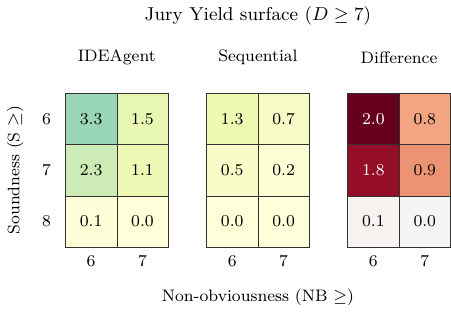}
        \caption{Zoomed in}
        \label{fig:yield-surfaces}
    \end{subfigure}\hfill
    \begin{subfigure}[b]{0.585\linewidth}
        \centering
        \includegraphics[width=\linewidth, trim=0 0 0 17pt, clip]{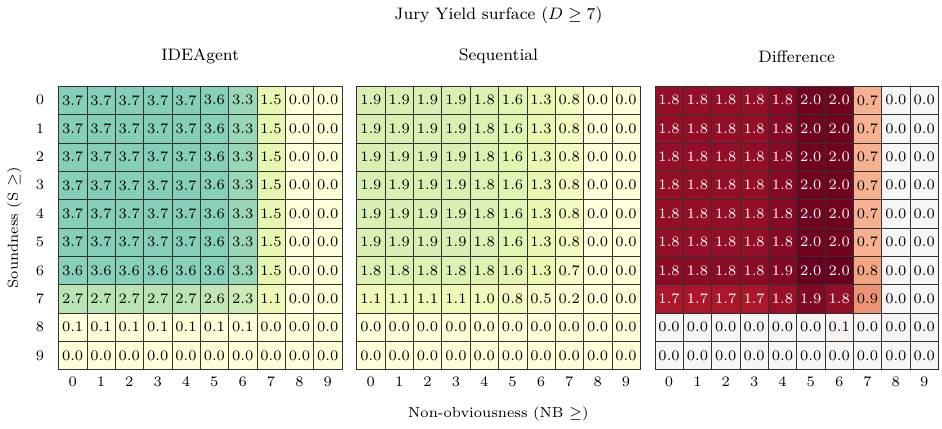}
        \caption{Full $10\times10$}
        \label{fig:yield-surface-full}
    \end{subfigure}
    \caption{\textit{Jury} averaged \yield surface ($\DIV \geq 7$) for \framework, Sequential-Memory baseline, and their difference.}
    \label{fig:yield-surfaces-both}
\end{figure}

To directly evaluate our core hypothesis that \framework produces a higher density of high-quality ideas within a diverse set, we compare the \yield of its generated ideas against Sequential-Memory. We define a \yield surface as a 2D grid in which each cell represents the \yield (out of 10) at a fixed diversity and clarity threshold while varying the soundness and non-obviousness thresholds. \Cref{fig:yield-surfaces} presents a zoomed-in view of the high-quality region ($NB \geq 6$, $S \geq 6$), along with the full $10 \times 10$ matrix in \Cref{fig:yield-surface-full}. The full grid reveals a broad \yield plateau for \framework under relaxed quality thresholds (top-left), where the \yield is $3.6$ for \framework compared to $1.9$ for Sequential-Memory, representing an improvement of $1.89\times$, at $(C \geq 6, D_{ij} \geq 7)$. As expected, the surface declines toward the bottom-right as the quality constraints become more stringent, eventually approaching zero for both methods when $NB \geq 8$ and $S \geq 8$.

The primary difference between the two setups becomes evident in the zoomed-in region ($NB \in [6,7]$, $S \in [6,8]$). Here, as the quality thresholds become stricter, the \yield decreases from $3.2$ (at $NB \geq 6$, $S \geq 6$) to $1.1$ (at $NB \geq 7$, $S \geq 7$), before eventually collapsing to zero. At the moderate quality threshold ($NB \geq 6$, $S \geq 6$), \framework achieves its largest improvement over the baseline, with a gain of $+1.9$. Even when the quality requirements are tightened to ($NB \geq 7$, $S \geq 7$), the gap remains substantial at $+0.9$, demonstrating its ability to generate non-obvious, sound, and mutually diverse research ideas. The advantage is also present under each evaluator separately as \framework has the highest mean \yield under both \opusold{} and \sonnet{} at $NB\ge6$ and $NB\ge7$ (\Cref{tab:standalone-results-all-baselines}). Against Sequential Memory, the topic-bootstrap 95\% confidence intervals for the jury-averaged mean differences are $[1.375,2.188]$ at $NB\ge6$ and $[0.656,1.156]$ at $NB\ge7$.

\paragraph{\framework vs. Baselines}
\framework also outperforms the secondary baseline, NOVA, as shown in \Cref{tab:main-standalone-results-all-baselines}, with per-judge results reported in the Appendix. The improvement in \yield is a clear $2\times$ and $3.89\times$ at $NB \geq 6$ and $NB \geq 7$, respectively. Furthermore, the three primary quality metrics all achieve their highest scores under \framework. The highest pairwise diversity is naturally achieved by NOVA owing to its final critic-based selection process, which explicitly optimizes for distinctness among the $3B$ candiates. 

This is followed by the One-Shot baseline, where all ideas share a common thinking space during generation, unlike the sequential approaches, which rely on a compact memory representation that is inherently susceptible to some information loss. However, over longer runs, compact signatures are substantially more computationally efficient for both comparison and context accumulation. Interestingly, One-Shot achieves a \yield of zero, indicating that none of its generated ideas satisfy the required thresholds. This suggests that maintaining a fully shared memory during generation can also be detrimental, as a poor or low-quality idea may directly influence subsequent generations and propagate its weaknesses. We therefore hypothesize that a lightweight sequential core-memory mechanism is sufficient to induce meaningful diversity while avoiding the risks of excessive cross-contamination between ideas. For the proportion of successful topics, \Cref{tab:main-standalone-results-all-baselines} shows that \framework has a clear edge over all baselines, indicating that it can produce qualifying ideas across a broader range of topics and quality--diversity gates. At $\DIV\!\ge\!7$, $\SO\!\ge\!7$, $\CL\!\ge\!6$, and $\NB\!\ge\!6$, the best baseline produces at least one qualifying idea on 25 of 32 topics, compared with 31 of 32 for \framework. When the non-obviousness threshold is increased to $\NB\!\ge\!7$, this difference becomes even sharper: the best baseline succeeds on 8 of 32 topics, whereas \framework succeeds on 27 of 32. Moreover, \Cref{tab:main-standalone-results-all-baselines} shows that the baselines attain substantially lower coverage when success requires multiple qualifying ideas. For example, at $Y_{6,7,6,7}$, NOVA produces at least two diverse, high-quality ideas on only 11 of 32 topics, whereas \framework does so on 23 of 32. The same pattern holds across the other reported quality-diversity gates. Taken together, these results reveal two complementary limitations of the strongest baselines: their qualifying outputs are often limited to a single idea rather than multiple diverse, high-quality ideas for the same topic, and their topic coverage falls sharply under the stricter non-obviousness threshold.

\paragraph{Quality improvements due to repair and refinement}

The primary distinction between the baselines and \framework is the inclusion of quality-improvement subroutines in the form of repair and refinement, which we analyze across all 320 lineages. \Cref{tab:quality-changes} illustrates the resulting quality improvements as assessed by the internal judge agents during generation. Specifically, we compare the initial seed idea with its final version after repair or refinement. Of the 320 lineages, only 30 required repair after falling just below the qualification thresholds, of which 28 were successfully improved and ultimately \textit{qualified}. Similarly, 182 of the 320 lineages underwent refinement after qualifying, and in a substantial $83\%$ of cases, the refined version replaced the original idea, showing that the refined version was selected in most cases.

\input{content_v2/tables/quality_and_agreement}
Taking a closer look, we analyze improvements across the individual quality rubrics together with the resulting \yield. Since refinement is applied only to ideas that have already qualified, whereas repair is reserved for ideas that narrowly miss the thresholds, refinement candidates naturally begin with higher absolute scores. Soundness benefits the most from both repair and refinement, reaching nearly $100\%$ qualification in both cases, with repair yielding a maximum improvement of $+23.2$. This suggests that targeted corrective feedback is highly effective at resolving logical inconsistencies. Clarity exhibits the second-largest improvement from repair ($+16.7$), while refinement improves it the most relative to the other rubrics, yielding an average gain of $+8.9$. Importantly, these improvements in soundness and clarity do not come at the expense of non-obviousness. Instead, both repair and refinement improve non-obviousness by approximately $+7$ points, indicating that gains across the different quality dimensions are complementary rather than competitive.

An interesting observation is that the final ideas produced by either process converge to remarkably similar scores across all quality rubrics. We hypothesize that this reflects an upper bound on the improvements achievable through targeted textual feedback, with diminishing returns after multiple refinement iterations. Finally, when analyzing \yield at the chosen thresholds, repair provides gains of $+0.88$ and $+0.87$ at the two non-obviousness thresholds, respectively, whereas refinement yields comparatively smaller gains.

It is also important to verify that the improvements observed by the internal agents during generation are reflected in downstream evaluation. \Cref{tab:quality-changes} also presents the corresponding analysis using the external judges. Although both sets of judges consistently agree on the direction of the improvements, the absolute gains differ substantially because of differences in their scoring scales. Clarity exhibits the largest downstream improvement, with gains of $+1.63$ and $+1.20$ points from repair and refinement, respectively, suggesting that improvements in clarity are readily recognized by independent evaluators. In contrast to the internal assessments during generation, the external judges reverse the relative effectiveness of repair and refinement. Refinement contributes the largest gains in \yield, improving it by $+0.88$ and $+0.78$ at the $\NB$ thresholds of $6$ and $7$, respectively. Finally, we refrain from directly correlating score deltas between the internal and external evaluations because the two systems use different rubrics and scoring scales. Instead, we emphasize their qualitative agreement regarding the consistent persistence of the observed improvements.

To establish the reliability of our evaluation framework, we analyze the
inter-annotator agreement between the two judges. As shown in
\Cref{tab:judge-agreement}, agreement per rubric is computed using
\textit{linear-weighted} Cohen's Kappa on the pooled evaluations across all
topics and runs. Across the primary rubrics, soundness shows the lowest
agreement ($0.26$), demonstrating that judging logical and mathematical
consistency of mechanisms and assumptions remains subjective
\citep{ho2026soundnessbenchaiscientistreally}. Clarity achieves the highest
agreement ($0.60$), indicating that clear ideas are readily identified by the
models. The remaining core rubrics, Non-obviousness and Diversity, exhibit
moderate agreement of $\sim0.43$, providing a reasonable indicator of the
jury's internal alignment. Taken together, agreement is consistently higher
for rubrics with concrete, checkable referents in an idea's text (clarity,
non-obviousness, diversity) than for those requiring more holistic or
forward-looking judgment (feasibility, significance, and to a lesser extent,
soundness), suggesting that judge disagreement reflects each rubric's inherent
ambiguity rather than a general unreliability of the evaluation framework. The Spearman column in \Cref{tab:judge-agreement} reports whether the two
evaluators rank the same ideas relatively higher or lower. Rank correlations
are positive across all rubrics, ranging from $0.409$ for soundness to $0.761$
for feasibility. According to \citet{schober2018correlation}, these values
indicate moderate rank consistency for non-obviousness, soundness, mechanism
clarity, significance, and pairwise diversity, and strong rank consistency for
feasibility. Interestingly, feasibility has low absolute-score agreement
($\kappa=0.210$) but strong rank consistency ($\rho=0.761$), indicating that
the judges use different score levels while largely agreeing on which ideas are
relatively more or less feasible.

%% file: content_v2/tables/headline_table_main.tex
\begin{table*}[t]
\centering
\caption{\textit{Jury}-averaged results across all 32 topics. Per-judge scores are provided in the Appendix. Bold marks the best method in each row. \textit{Note:} $^{\dagger}p<.05$; $^{\ddagger}p<.001$; $^{\mathsection}p<.0001$ (paired topic-level tests; rubric tests Holm-corrected).}
\label{tab:main-standalone-results-all-baselines}
\renewcommand{\arraystretch}{1.15}
\setlength{\tabcolsep}{6pt}
\small
\begin{tabularx}{\textwidth}{@{}X*{5}{>{\centering\arraybackslash}X}@{}}
\toprule
Metric & $\mathcal{SM}$ & $\mathcal{OS}$ & Stateless & NOVA & \framework \\
\midrule
\rowcolor{gray!15}
\multicolumn{6}{l}{\textsc{QD outcomes}} \\
\quad $Y_{7,7,6,7}$ &
$0.188$ &
$0.000^{\dagger}$ &
$0.250$ &
$0.281$ &
$\mathbf{1.094}^{\mathsection}$ \\
\quad $Y_{6,7,6,7}$ &
$0.531$ &
$0.000^{\ddagger}$ &
$0.812^{\dagger}$ &
$1.156^{\dagger}$ &
$\mathbf{2.312}^{\mathsection}$ \\
\midrule
\addlinespace[3pt]
\rowcolor{blue!10}
\multicolumn{6}{l}{\textsc{Successful topics}} \\
\quad $Y_{7,7,6,7}\!\ge\!1$ & $6/32$ & $0/32$ & $7/32$ & $8/32$ & $\mathbf{27/32}$ \\
\quad $Y_{7,7,6,7}\!\ge\!2$ & $0/32$ & $0/32$ & $1/32$ & $1/32$ & $\mathbf{8/32}$ \\
\quad $Y_{6,7,6,7}\!\ge\!1$ & $14/32$ & $0/32$ & $24/32$ & $25/32$ & $\mathbf{31/32}$ \\
\quad $Y_{6,7,6,7}\!\ge\!2$ & $3/32$ & $0/32$ & $2/32$ & $11/32$ & $\mathbf{23/32}$ \\
\quad $Y_{6,7,6,7}\!\ge\!3$ & $0/32$ & $0/32$ & $0/32$ & $1/32$ & $\mathbf{15/32}$ \\
\midrule
\addlinespace[3pt]
\rowcolor{green!10}
\multicolumn{6}{l}{\textsc{Individual dimensions}} \\
\quad NB &
$6.020$ & $5.609^{\mathsection}$ & $6.077$ & $6.142$ & $\mathbf{6.430}^{\mathsection}$ \\
\quad Soundness &
$6.534$ & $5.477^{\mathsection}$ & $6.589$ & $6.483$ & $\mathbf{6.720}^{\dagger}$ \\
\quad Clarity &
$5.416$ & $4.030^{\mathsection}$ & $5.739^{\dagger}$ & $5.556$ & $\mathbf{6.341}^{\mathsection}$ \\
\quad Diversity &
$6.606$ & $6.739^{\dagger}$ & $5.080^{\ddagger}$ & $\mathbf{6.757}^{\dagger}$ & $6.641$ \\
\quad Feasibility &
$4.053$ & $4.034$ & $4.006$ & $\mathbf{4.150}$ & $3.864$ \\
\quad Significance &
$5.683$ & $5.316^{\mathsection}$ & $\mathbf{5.898}^{\dagger}$ & $5.470^{\dagger}$ & $5.572$ \\
\bottomrule
\end{tabularx}
\end{table*}

%% file: content_v2/tables/quality_and_agreement.tex
\begin{table}[t]
\centering
\footnotesize
\setlength{\tabcolsep}{4.5pt}
\caption{Quality improvements from repair and refinement.}
\label{tab:quality-changes}
\begin{tabular}{@{}ll ccccc @{\hspace{12pt}} ccccc@{}}
\toprule
 & & \multicolumn{5}{c}{\textsc{Internal Judges}} & \multicolumn{5}{c}{\textsc{External Jury}} \\
\cmidrule(lr){3-7}\cmidrule(l){8-12}
 & & \multicolumn{2}{c}{\textbf{Score}} & \multicolumn{3}{c}{\textbf{Lineages} (\%)} & \multicolumn{2}{c}{\textbf{Score}} & \multicolumn{3}{c}{\textbf{Lineages} (\%)} \\
\cmidrule(lr){3-4}\cmidrule(lr){5-7}\cmidrule(lr){8-9}\cmidrule(l){10-12}
 & & Seed & Final & $\uparrow$ & $=$ & $\downarrow$ & Seed & Final & $\uparrow$ & $=$ & $\downarrow$ \\
\midrule
\multirow{5}{*}{\textbf{Repair}}
 & $\NB$     & $85.3$ & $91.7$ & $64.3$ & $28.6$ & $7.1$ & $6.44$ & $6.62$ & $35.7$ & $39.3$ & $25.0$ \\
 & Soundness & $76.4$ & $99.6$ & $96.4$ & $3.6$  & $0.0$ & $6.39$ & $6.88$ & $53.6$ & $35.7$ & $10.7$ \\
 & Clarity   & $76.4$ & $93.1$ & $78.6$ & $14.3$ & $7.1$ & $5.29$ & $6.92$ & $85.7$ & $14.3$ & $0.0$  \\
 & $Y_{60,60,60,60}$ / $Y_{6,7,6,7}$ & $8.53$ & $9.41$ & \multicolumn{3}{c}{--} & $1.12$ & $1.38$ & \multicolumn{3}{c}{--} \\
 & $Y_{70,60,60,60}$ / $Y_{7,7,6,7}$ & $8.44$ & $9.31$ & \multicolumn{3}{c}{--} & $0.19$ & $0.34$ & \multicolumn{3}{c}{--} \\
\midrule
\multirow{5}{*}{\textbf{Refinement}}
 & $\NB$     & $83.9$ & $91.3$ & $86.0$ & $13.3$ & $0.7$ & $6.08$ & $6.48$ & $58.0$ & $31.5$ & $10.5$ \\
 & Soundness & $94.3$ & $99.4$ & $67.8$ & $28.0$ & $4.2$ & $6.62$ & $6.81$ & $51.0$ & $34.3$ & $14.7$ \\
 & Clarity   & $84.5$ & $93.4$ & $80.4$ & $16.8$ & $2.8$ & $5.70$ & $6.90$ & $83.9$ & $11.9$ & $4.2$  \\
 & $Y_{60,60,60,60}$ / $Y_{6,7,6,7}$ & $8.53$ & $8.53$ & \multicolumn{3}{c}{--} & $1.12$ & $2.00$ & \multicolumn{3}{c}{--} \\
 & $Y_{70,60,60,60}$ / $Y_{7,7,6,7}$ & $8.44$ & $8.53$ & \multicolumn{3}{c}{--} & $0.19$ & $0.97$ & \multicolumn{3}{c}{--} \\
\bottomrule
\end{tabular}
\end{table}

\begin{table}[t]
\centering
\footnotesize
\setlength{\tabcolsep}{8pt}
\caption{Agreement between \opusold{} and \sonnet{} across all methods and topics.}
\label{tab:judge-agreement}
\begin{tabular}{lcc}
\toprule
Rubric & Linear-weighted $\kappa$ & Median Spearman $\rho$ \\
\midrule
NB                 & $0.455$ & $0.610$ \\
Soundness          & $0.268$ & $0.409$ \\
Clarity            & $\mathbf{0.604}$ & $0.600$ \\
Feasibility        & $0.210$ & $\mathbf{0.761}$ \\
Significance       & $0.352$ & $0.504$ \\
Diversity          & $0.424$ & $0.679$ \\
\bottomrule
\end{tabular}
\end{table}

%% file: content_v2/verifiable.tex
\section{Extension to Verifiable Algorithm Discovery}
\paragraph{Setup} Most of the experiments and analysis in this work were conducted for open-domain ideation, where a \textit{holistic} quality evaluation is tedious and most relied on automated subjective evaluation with LLMs, which we attempt to corroborate with human evaluation in Appendix \ref{app:results}. 

To study the efficacy of \framework with minimal reliance on LLMs for evaluation, we also run the ideation in a verifiable setup where the \textit{quality} or performance of ideas can be evaluated automatically via a standard metric. To this end, we choose \alpharesearch \citep{yu2026alpharesearchacceleratingnewalgorithm} and \alebench \citep{imajuku2025alebench}, which have program-based tasks spanning various domains such as algorithm design, data structure design, heuristic search, and combinatorial optimization, along with a \textit{best reference} for the task. Analogous to real-world research, we add a trivial/standard baseline for each task that our search is expected to cross. \yield is retained as the QD metric, where ideas below a set quality threshold are filtered, and the largest set of pairwise distinct ideas is extracted. The exact adaptation details are documented in Appendix \ref{app:alpharesearch}.

\paragraph{Why these benchmarks support a QD search} Both benchmarks fix the \textit{problem} and its \textit{objective} but leave the \textit{method} open, which is the setting a QD search is designed for: every task admits many valid solutions of different quality, and no single approach is known to be optimal, hence both \textit{how good an idea is} and \textit{how different it is from the others} are meaningful. The quality of an idea is also verifiable, since its program is executed and its output is scored by the benchmark's own metric. This replaces the judged quality axis of the open domain with a measured one, and the trivial solution and the external reference\footnote{\textit{human-best} on \alpharesearch, the \textit{contest-winner} on \alebench} anchor every score on a common scale from which the \yield thresholds are drawn. Further, the two benchmarks cover complementary regimes. \alpharesearch poses mathematical construction problems with a single instance per task, where several unrelated construction and search principles can succeed, so diversity is cheap and quality rewards depth. \alebench poses heuristic optimization problems from competitive programming, where a program must perform well across many unseen test cases within a strict time limit, against strong human-engineered solutions, so a few standard techniques are hard to beat, and finding several distinct methods that clear the bar is considerably harder. Together, they test whether the search works when diverse solutions are plentiful and scarce.

\input{content_v2/tables/verifiable_yield}

\paragraph{Results} From the \yield scores presented in \Cref{tab:verifiable-clique-ensemble}, we find that \framework consistently outperforms all the baselines on both benchmarks. Its best idea per problem reaches $Q_{\max} = 106.1$ on \alpharesearch and $56.4$ on \alebench, $12.8$ and $11.3$ points above the strongest baseline, and its average idea is also the strongest on \alpharesearch, while on \alebench it matches Stateless. Thus, \framework raises the quality ceiling without lowering the average, and does not trade quality for diversity. 

Apart from this, the advantage of \framework grows as the quality bar rises, i.e., on \alpharesearch, it finds $1.24\times$ as many distinct ideas as the best baseline among all those that clear the trivial solution, but $1.75\times$ as many at $Q_i \ge 75$, and on \alebench the margin doubles at $Q_i \ge 75$. The ideas that make \framework diverse are therefore also its strong ones, whereas the distinct ideas of the baselines are concentrated at low quality. Secondly, no baseline is consistently the runner-up. NOVA, whose final selection explicitly favors distinctness, has the most distinct ideas among the baselines on \alpharesearch at $Y_0$, but falls behind as the threshold rises, while Stateless, which repeatedly samples the model's strongest instinct, is the best baseline at $Q_i \ge 75$ but trails at $Y_0$. Each baseline thus trades one axis for the other, and only \framework leads at every threshold on both benchmarks. Notably, \framework finds $0.67$ distinct ideas per task that reach the \textit{human-best} on \alpharesearch, against at most $0.56$ for any baseline.

%% file: content_v2/tables/verifiable_yield.tex
\begin{table}[t]
\centering
\small
\caption{\yield of the verifiable searches under the three-judge jury. $Y_k$ is the size of the largest set of mutually distinct ideas among those that clear the trivial solution with progress $Q_i \ge k$, and $Y_0$ counts every idea that clears it. $Q_i = 100$ is the \textit{human-best} and the \textit{contest-winner}, respectively. $Q_{\max}$ and $Q_{\text{avg}}$ are the highest and the mean $Q_i$ among the ten final ideas of a problem. Benchmark scores are averaged over the number of problems in it, and bold marks the best method in each column.}
\label{tab:verifiable-yield}\label{tab:verifiable-clique-ensemble}
\setlength{\tabcolsep}{4.5pt}
\renewcommand{\arraystretch}{1.1}
\begin{tabular}{@{}l cc@{\hspace{12pt}}cccc @{\hspace{16pt}} cc@{\hspace{12pt}}cccc@{}}
\toprule
 & \multicolumn{6}{c}{\alpharesearch} & \multicolumn{6}{c}{\alebench} \\
\cmidrule(lr){2-7} \cmidrule(l){8-13}
Method & $Q_{\max}$ & $Q_{\text{avg}}$ & $Y_0$ & $Y_{50}$ & $Y_{75}$ & $Y_{100}$ & $Q_{\max}$ & $Q_{\text{avg}}$ & $Y_0$ & $Y_{25}$ & $Y_{50}$ & $Y_{75}$ \\
\midrule
$\mathcal{SM}$ & 92.6 & 46.1 & 4.00 & 3.11 & 1.67 & 0.56 & 38.3 & 18.6 & 2.29 & 1.71 & 1.14 & 0.29 \\
$\mathcal{OS}$ & 53.2 & 16.9 & 2.78 & 1.00 & 0.44 & 0.33 & 14.8 & 6.2 & 1.14 & 0.43 & 0.43 & 0.00 \\
Stateless      & 93.5 & 47.8 & 4.11 & 3.11 & 1.78 & 0.56 & 45.1 & \textbf{22.7} & 2.29 & 1.71 & 0.57 & 0.00 \\
NOVA           & 75.4 & 34.4 & 4.56 & 2.67 & 0.78 & 0.22 & 37.3 & 18.0 & 2.29 & 2.00 & 1.00 & 0.00 \\
\framework     & \textbf{106.2} & \textbf{54.5} & \textbf{5.67} & \textbf{4.44} & \textbf{3.11} & \textbf{0.67} & \textbf{56.4} & 22.6 & \textbf{3.43} & \textbf{2.57} & \textbf{1.57} & \textbf{0.57} \\
\bottomrule
\end{tabular}
\end{table}

%% file: content_v2/conclusion.tex
\section{Conclusion}
\label{sec:conclusion}
In this work, we presented \framework, a multi-agent framework that reframes scientific ideation as a QD search to build \textit{portfolios of ideas} with high quality density. Through sequential generation and multi-agent evaluation centered on the core requirements of QD search, we introduced a quality-improvement subroutine that repairs or optionally refines ideas based on identified \textit{opportunities}. We further demonstrated lineage-inspired idea management using dedicated archives for completed, rejected, and historical ideas, efficiently enabled by core signature comparison. To jointly evaluate quality and diversity, we proposed \yield, which filters ideas using fixed thresholds on its primary quality indicators before selecting the largest subset whose ideas satisfy a pairwise diversity threshold. Finally, we evaluated \framework across 32 Computer Science research topics and demonstrated the contributions of both the quality-improvement loop and compact stateful memory. Its extension to verifiable algorithm discovery also holds as the best while relaxing subjective evaluations from LLM-judges. Lastly, we hope our problem formulation provides a new perspective on AI-driven scientific ideation.

%% file: content_v2/extra.tex
\subsection*{AI use statement}
In this work, we have used AI for code writing and repository management. All the code written by AI was checked and verified to run correctly and as required by two authors of this paper. Additionally, we used AI for initial brainstorming, i.e., for quick idea testing and verification, and not full research problem generation. Final draft grammatical polishing and rephrasing were done with the help of AI, but the main content was still written by the authors. All the related work and research gaps were identified by the authors themselves without the use of AI anywhere. Finally, we take responsibility for the final content of this work, including text, claims, or artifacts produced with the aid of generative AI.

\subsection*{Ethics statement}
In this work, we study and develop an agentic framework for automated idea generation and scientific discovery. However, the ideas generated might be incomplete, unsafe, illegal, impractical, or overlapping with existing works. Hence, we strongly argue and resonate with the fact that all the AI-generated ideas should be proofread and verified by humans before working on them further in a \textbf{practical deployment}.

\subsection*{Reproducibility statement}
Our entire codebase was developed \textbf{exactly} around the setup described in the paper. Further, we plan to completely open-source it, along with the background data used for idea generation. All the papers used from the background data collection and verifiable benchmarking were publicly available on the web, and directly scraped with no private modifications. The Agents used in the paper are also publicly available as APIs from the respective providers, or on \texttt{HuggingFace}. However, the exact generation outputs might vary slightly depending on the reasoning effort and sampling strategies used by them internally, and we do not have any control over it. All the data processing and main methodology pieces are well-described in the main paper, and no significant section is withheld. The exact algorithms used for topic selection and \yield are also described as is in the appendix. 

\subsection*{Acknowledgements}
We extend our sincere gratitude to Google DeepMind for their \textit{Gemini Academic Program Award} and GCP credits, which enabled us to run the Gemini and Claude models at scale for our project. We also thank Nithish Selvaraj for his inputs and suggestions on the verifiable domain setup.

%% file: content_v2/appendix/limitations.tex
\section{Limitations and Scope of Work}
\label{app:limitations}

Our work on open-ended idea generation has several limitations that warrant discussion.

\begin{enumerate}
\item All internal and external evaluations rely primarily on LLM judges to assess non-obviousness, soundness, clarity, feasibility, and diversity. To mitigate evaluator-specific bias, we employ two independent large-scale judge models and average their scores. Likewise, to calibrate the crucial soundness evaluation, we aggregate five independently sampled scores to mitigate sampling bias. Moreover, our primary metric, \yield, is model-agnostic and can be paired with alternative evaluation protocols, including human judgments.

\item We restrict this work to generating portfolios of diverse, high-quality research ideas and evaluating them accordingly. Consequently, we do not evaluate immediate \textbf{real-world} feasibility, exact correctness, or practical effectiveness of the generated ideas, which would require implementation, debugging, end-to-end experimentation, and expert validation. Likewise, we \textbf{do not} claim that the generated ideas in our experiments are necessarily immediately publishable or absent from the existing literature. Verifying such claims would require a comprehensive retrieval over both published and emerging research to identify related prior work, along with a significantly increased monetary investment beyond our scope. However, we do show that our ideas exhibit strong soundness and logical consistency across multiple rigorous evaluation. 

\item Each idea is given one repair opportunity and at most two refinements, if one is not consumed by repair. Both repair and refinement are computationally expensive because they require all agents (Quality, Diversity, Soundness, Ideator, and Critic). Since our setup uses proprietary LLMs, additional rounds would substantially increase monetary cost. More generally, we do not rigorously optimize hyperparameters (rubric thresholds, iteration counts, and API generation settings) because of computational, monetary, and time constraints. Consequently, the reported results may be under-optimized. Nevertheless, our goal is to demonstrate the value of repair and refinement for improving ideas along the Quality axis.

\item All experiments use proprietary LLMs accessed through their APIs because we found that \textbf{current} smaller open-source models lacked sufficient knowledge, idea formulation/refinement, and evaluation capability. However, our framework is model-agnostic, and we expect the underlying principles of sequential generation, targeted multi-dimensional evaluation, and feedback to transfer across future models. In preliminary experiments, we evaluated \texttt{Gemma-4-31B}, \texttt{Qwen-3.6-27B}, \texttt{GPT-OSS-120B}, \geminipro{}, \sonnet{}, and \opus{}. These models generally produced non-obvious, diverse ideas with clearly articulated mechanisms; however, the evaluators \textit{frequently} identified serious soundness flaws, substantially reducing overall quality. \fable{} generated moderately sound, highly non-obvious, and diverse ideas, but budget constraints precluded its inclusion in the main experiments. In a pilot run on \texttt{cs.AI-001}, the internal judge assigned its ideas a mean soundness score of $66.4$, compared with $91.8$ for \gpt{}, triggering several repairs of the initial seed drafts. Relative to \gpt{}, \fable{} achieved a $9$-point advantage in non-obviousness but scored three points lower in diversity. Thus, despite its novelty advantage, its lower initial soundness increased the required repair effort and, together with the budget constraints, motivated our decision not to use it as the primary Ideator.
\item Due to the authors belonging to the CS and AI domain, we restrict the data collection and experiments to them. But we strongly suggest that the proposed method is generic and can also be transposed to other sciences such as Chemistry, Biology, or Physics. 
\item Most importantly, the motivation of this work is \textbf{not} to claim SoTA in any manner, but just present a \textbf{proof-of-concept} for the new QD formulation of the ideation process.
\end{enumerate}

%% file: content_v2/appendix/extra_results.tex
\section{Additional Results and Discussions}
\label{app:results}

\input{content_v2/tables/per_judge_appendix}

\input{content_v2/tables/jury_verdict_contingency}
\paragraph{Human evaluation.} As a small comparative check against human ratings, two authors scored 25 topic-matched idea pairs on the same three quality rubrics used by the Claude jury, blind to which system produced each idea. Table~\ref{tab:human-eval} shows that human raters independently reproduce the automatic ranking. \framework is preferred on $69.6\%$ of topic-matched pairs ($p<.05$), and leads on all three rubrics, and on no rubric does NOVA lead under either jury. The magnitude of the advantage is also consistent across juries: the human mean margin ($+0.47$) closely matches the Claude jury's ($+0.52$), despite the two juries operating on different absolute scales, with human raters scoring roughly $0.7$ points lower overall.

Table~\ref{tab:jury-verdict-contingency} reports the joint distribution of per-topic verdicts. The juries jointly favour \framework on 12 of 25 topics and jointly favour NOVA on only one. Where they differ, the disagreement is clearly one-sided, i.e., on $6$ topics the human jury awards the pair to NOVA while the Claude jury awards it to \framework, against $2$ in the opposite direction. Hence, the human jury is the more conservative of the two, returning a decisive verdict for \framework on 16 topics against the Claude jury's 20, rather than being systematically opposed to it. Both juries converge on the \textit{same overall conclusion} from independent evidence, with the human study providing the stricter test.

We note that this study uses two raters over 25 pairs and is aimed to test the overall ranking rather than per-rubric differences. The individual rubric margins are directionally consistent but do not reach significance under the human jury.

\input{content_v2/tables/human_eval}

%% file: content_v2/tables/per_judge_appendix.tex
\begin{table*}[t]
\centering
\caption{
Results across 32 topics as evaluated by the two external judges.
Bold marks the best method in each row.
\textit{Note.} $^{\dagger}p<.05$;
$^{\ddagger}p<.001$;
$^{\mathsection}p<.0001$
(paired topic-level tests; rubric tests Holm-corrected).
}
\label{tab:standalone-results-all-baselines}
\renewcommand{\arraystretch}{1.30}
\setlength{\tabcolsep}{3.25pt}
\begin{subtable}[t]{0.495\textwidth}
\centering
\vspace{0pt}
\small
\resizebox{\linewidth}{!}{%
\begin{tabular}{lccccc}
\toprule
Metric & $\mathcal{SM}$ & $\mathcal{OS}$ & Stateless & NOVA & \framework \\
\midrule
\rowcolor{gray!15}
\multicolumn{6}{l}{\textsc{QD outcomes}} \\[-1pt]
\quad $Y_{7,7,6,7}$ &
$1.312$ &
$0.031^{\mathsection}$ &
$1.312$ &
$1.562$ &
$\mathbf{3.531}^{\mathsection}$ \\
\quad $Y_{6,7,6,7}$ &
$2.375$ &
$0.031^{\mathsection}$ &
$1.719^{\dagger}$ &
$2.844$ &
$\mathbf{4.656}^{\mathsection}$ \\
\midrule
\addlinespace[3pt]
\rowcolor{blue!10}
\multicolumn{6}{l}{\textsc{Successful topics}} \\[-1pt]
\quad $Y_{7,7,6,7}\!\ge\!1$ &
$22/32$ & $1/32$ & $28/32$ & $23/32$ & $\mathbf{32/32}$ \\
\quad $Y_{7,7,6,7}\!\ge\!2$ &
$14/32$ & $0/32$ & $12/32$ & $17/32$ & $\mathbf{31/32}$ \\
\quad $Y_{7,7,6,7}\!\ge\!3$ &
$6/32$ & $0/32$ & $2/32$ & $7/32$ & $\mathbf{26/32}$ \\
\quad $Y_{6,7,6,7}\!\ge\!1$ &
$27/32$ & $1/32$ & $31/32$ & $31/32$ & $\mathbf{32/32}$ \\
\quad $Y_{6,7,6,7}\!\ge\!2$ &
$21/32$ & $0/32$ & $19/32$ & $27/32$ & $\mathbf{31/32}$ \\
\quad $Y_{6,7,6,7}\!\ge\!3$ &
$17/32$ & $0/32$ & $5/32$ & $18/32$ & $\mathbf{30/32}$ \\
\midrule
\addlinespace[3pt]
\rowcolor{green!12}
\multicolumn{6}{l}{\textsc{Individual dimensions}} \\[-1pt]
\quad NB &
$6.175$ &
$5.609^{\mathsection}$ &
$6.206$ &
$6.331$ &
$\mathbf{6.612}^{\mathsection}$ \\
\quad Soundness &
$6.862$ &
$5.794^{\mathsection}$ &
$6.891$ &
$6.834$ &
$\mathbf{6.978}$ \\
\quad Clarity &
$5.447$ &
$3.878^{\mathsection}$ &
$5.728$ &
$5.597$ &
$\mathbf{6.353}^{\mathsection}$ \\
\quad Diversity &
$7.062$ &
$7.190^{\dagger}$ &
$5.309^{\mathsection}$ &
$\mathbf{7.268}^{\dagger}$ &
$7.047$ \\
\quad Feasibility &
$4.862$ &
$4.772$ &
$4.847$ &
$\mathbf{5.006}$ &
$4.637$ \\
\quad Significance &
$5.975$ &
$5.534^{\mathsection}$ &
$\mathbf{6.137}$ &
$5.775^{\dagger}$ &
$5.822^{\dagger}$ \\
\bottomrule
\end{tabular}%
}
\caption{\opusold{}}
\label{tab:opus-standalone-results-all-baselines}
\end{subtable}
\hfill
\begin{subtable}[t]{0.495\textwidth}
\centering
\vspace{0pt}
\small
\resizebox{\linewidth}{!}{%
\begin{tabular}{lccccc}
\toprule
Metric & $\mathcal{SM}$ & $\mathcal{OS}$ & Stateless & NOVA & \framework \\
\midrule
\rowcolor{gray!15}
\multicolumn{6}{l}{\textsc{QD outcomes}} \\[-1pt]
\quad $Y_{7,7,6,7}$ &
$0.406$ &
$0.000^{\ddagger}$ &
$0.375$ &
$0.500$ &
$\mathbf{1.219}^{\mathsection}$ \\
\quad $Y_{6,7,6,7}$ &
$0.844$ &
$0.031^{\mathsection}$ &
$1.000$ &
$1.250^{\dagger}$ &
$\mathbf{2.406}^{\mathsection}$ \\
\midrule
\addlinespace[3pt]
\rowcolor{blue!10}
\multicolumn{6}{l}{\textsc{Successful topics}} \\[-1pt]
\quad $Y_{7,7,6,7}\!\ge\!1$ &
$12/32$ & $0/32$ & $11/32$ & $14/32$ & $\mathbf{28/32}$ \\
\quad $Y_{7,7,6,7}\!\ge\!2$ &
$1/32$ & $0/32$ & $1/32$ & $2/32$ & $\mathbf{10/32}$ \\
\quad $Y_{6,7,6,7}\!\ge\!1$ &
$20/32$ & $1/32$ & $28/32$ & $28/32$ & $\mathbf{31/32}$ \\
\quad $Y_{6,7,6,7}\!\ge\!2$ &
$6/32$ & $0/32$ & $4/32$ & $11/32$ & $\mathbf{26/32}$ \\
\quad $Y_{6,7,6,7}\!\ge\!3$ &
$1/32$ & $0/32$ & $0/32$ & $1/32$ & $\mathbf{15/32}$ \\
\midrule
\addlinespace[3pt]
\rowcolor{green!12}
\multicolumn{6}{l}{\textsc{Individual dimensions}} \\[-1pt]
\quad NB &
$5.866$ &
$5.609^{\dagger}$ &
$5.947$ &
$5.953$ &
$\mathbf{6.247}^{\mathsection}$ \\
\quad Soundness &
$6.206$ &
$5.159^{\mathsection}$ &
$6.287$ &
$6.131$ &
$\mathbf{6.463}^{\ddagger}$ \\
\quad Clarity &
$5.384$ &
$4.181^{\mathsection}$ &
$5.750^{\dagger}$ &
$5.516$ &
$\mathbf{6.328}^{\mathsection}$ \\
\quad Diversity &
$6.150$ &
$\mathbf{6.288}$ &
$4.850^{\mathsection}$ &
$6.246$ &
$6.234$ \\
\quad Feasibility &
$3.244$ &
$\mathbf{3.297}$ &
$3.166$ &
$3.294$ &
$3.091$ \\
\quad Significance &
$5.391$ &
$5.097^{\ddagger}$ &
$\mathbf{5.659}^{\dagger}$ &
$5.166^{\dagger}$ &
$5.322$ \\
\bottomrule
\end{tabular}%
}
\caption{\sonnet{}}
\label{tab:sonnet-standalone-results-all-baselines}
\end{subtable}
\end{table*}

%% file: content_v2/tables/jury_verdict_contingency.tex
\begin{table}[t]
\centering
\footnotesize
\caption{Joint distribution of per-topic verdicts. Each jury returns one verdict per topic---\framework wins, tie, or NOVA wins---given by the sign of its mean \framework minus NOVA score difference over the three rubrics. Both juries independently favor \framework on the large majority of topics ($20/25$ and $16/25$), and jointly favor NOVA on a single topic.}
\label{tab:jury-verdict-contingency}
\begin{tabular}{@{}lcccc@{}}
\toprule
& \multicolumn{3}{c}{Claude jury} & \\
\cmidrule(lr){2-4}
Human jury & \framework & Tie & NOVA & Total \\
\midrule
\framework wins & $\mathbf{12}$ & $2$ & $2$ & $16$ \\
Tie & $2$ & $0$ & $0$ & $2$ \\
NOVA wins & $6$ & $0$ & $\mathbf{1}$ & $7$ \\
\midrule
Total & $20$ & $2$ & $3$ & $25$ \\
\bottomrule
\end{tabular}
\end{table}

%% file: content_v2/tables/human_eval.tex
\begin{table*}[ht]
\centering
\small
\caption{Human evaluation of \framework against Nova, the strongest baseline, alongside the Claude jury scoring the same ideas. Two expert raters scored 25 topic-matched idea pairs on a 0--9 rubric, blind to source. Win rate is the share of pairs on which \framework scores higher, excluding ties. Human values average both raters; Claude values average Claude Opus 4.7 and Claude Sonnet 5; Overall is the equal-weight mean of the three rubrics. $^{\dagger}p<.05$, $^{\ddagger}p<.001$ (Wilcoxon signed-rank over per-topic score differences).}
\label{tab:human-eval}
\begin{tabular}{@{}lcccccc@{}}
\toprule
& \multicolumn{3}{c}{Human jury} & \multicolumn{3}{c}{Claude jury} \\
\cmidrule(lr){2-4} \cmidrule(lr){5-7}
Rubric & \framework & NOVA & Win rate & \framework & NOVA & Win rate \\
\midrule
NB & $5.48$ & $5.10$ & $61.9\%$ & $6.44$ & $6.18$ & $63.2\%$ \\
Soundness & $5.68$ & $5.18$ & $62.5\%$ & $6.80$ & $6.32$ & $93.8\%^{\dagger}$ \\
Mechanism clarity & $6.38$ & $5.86$ & $70.0\%$ & $6.46$ & $5.64$ & $81.0\%^{\dagger}$ \\
\midrule
\textbf{Overall} & $\mathbf{5.85}$ & $5.38$ & $\mathbf{69.6\%}^{\dagger}$ & $\mathbf{6.57}$ & $6.05$ & $\mathbf{87.0\%}^{\ddagger}$ \\
\bottomrule
\end{tabular}
\end{table*}

%% file: content_v2/appendix/cost.tex
\section{Cost and Futher Scaling}
\label{app:cost}

\paragraph{Cost / Successful Topic}
A natural question in research ideation is whether it is better to sample more ideas using cheaper methods or fewer ideas using more capable but expensive ones. To answer this, we report the Cost per Successful Topic (CST) of a method $M$, defined as the total generation cost across all attempted topics divided by the number of successful topics:
\begin{equation}
    \operatorname{CST}_{M}(k,l,m,\tau,\Phi) = \frac{\sum_{T\in \text{Topics}} C_{M,T}} {\sum_{T\in \text{Topics}} \mathds{1}\!\left[Y^{M,T}_{k,l,m,\tau} \ge \Phi\right]}
\end{equation}
where $C_{M,T}$ is the generation cost of method $M$ on topic $T$, and $\Phi$ is the required yield. $n(\text{Topics})=32$, $(l,m,\tau)=(7,6,7)$, $k\in\{6,7\}$, and $\Phi\in\{1,2,3\}$. If no topic reaches the required yield, CST is reported as \textit{NA}.

The objective of QD search is to generate diverse, high-quality ideas for each topic. The pairwise diversity constraint is meaningful only when at least two ideas qualify; consequently, $\Phi=2$ is our primary diverse-portfolio criterion.

Stateless is the most cost-effective method when success requires only one qualifying idea, although this criterion does not exercise pairwise diversity. When success requires at least two diverse, high-quality ideas for the same topic, \framework becomes the most cost-effective method at both non-obviousness thresholds. Compared with the strongest baseline, it is approximately \(1.6\times\) cheaper per successful topic while succeeding on \(2.1\times\) as many topics at \(\NB\!\ge\!6\) and \(8\times\) as many topics at \(\NB\!\ge\!7\). At the stricter requirement of three qualifying ideas, \framework is approximately \(11\times\) cheaper than the only baseline that succeeds.

\input{content_v2/tables/cost_per_successful_topic}

\paragraph{Scaling the Baselines}
We scaled the Stateless and Sequential Memory baselines to approximately match \framework's generation budget. Stateless was scaled from 10 to 50 fresh generations across all 32 topics, while Sequential Memory was scaled from 10 to 30 and 45 generations on a matched subset of four topics to examine its scaling behaviour. From each generation pool, up to 10 ideas were retained by \geminipro{} using the selection criteria described above.

\begin{figure*}[t]
\centering
\includegraphics[width=\textwidth]{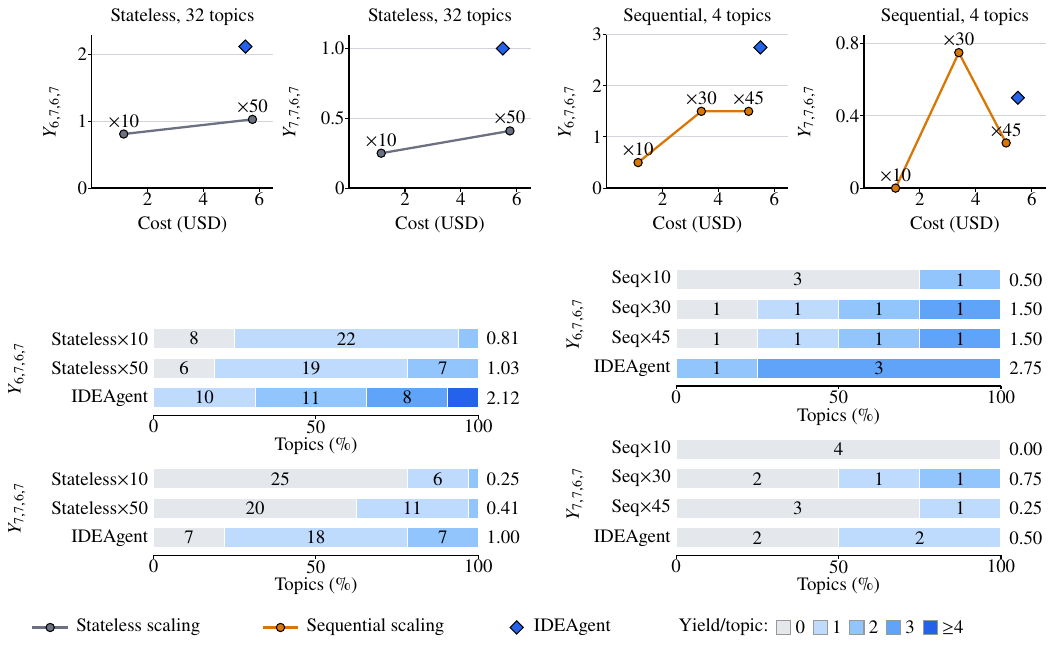}
\caption{Scaling fresh generation does not reproduce \framework's
quality--diversity performance. \textbf{Top:} cost--\yield at both quality
gates, for Stateless on 32 matched topics and Sequential-Memory on four.
\textbf{Bottom:} distribution of per-topic \yield; the number beside each bar
is that arm's mean.}
\label{fig:budget-scaling-combined}
\end{figure*}

As shown in \Cref{fig:budget-scaling-combined}, scaling Stateless from 10 to 50 generations increases mean Yield by approximately 27\% at the $\NB\ge6$ gate and 62\% at the $\NB\ge7$ gate. Nevertheless, \framework achieves approximately $2.06\times$ and $2.46\times$ higher Yield than the scaled Stateless baseline at the respective gates. The per-topic Yield distributions show the same trend: additional fresh generations improve Stateless but do not match \framework's ability to produce multiple qualifying ideas per topic.

On the four-topic subset, Sequential Memory improves threefold at $\NB\ge6$ when scaled from 10 to 30 generations, but shows no further gain at 45 generations. At $\NB\ge7$, performance is non-monotonic, increasing at 30 generations before declining at 45. Although this smaller experiment is exploratory, its per-topic Yield distribution likewise indicates that additional sampling alone does not reproduce \framework's performance. Together, these experiments suggest that \framework's gains cannot be explained solely by a larger generation budget.

%% file: content_v2/tables/cost_per_successful_topic.tex
\begin{table}[t]
\centering
\renewcommand{\arraystretch}{1.30}
\setlength{\tabcolsep}{4pt}
\small
\caption{Cost (USD) per successful topic across 32 topics, evaluated by two
external judges. Bold marks the cheapest method in each row; \textit{NA}
indicates the method never reached that yield threshold within budget.}
\label{tab:cost-per-successful-topic}
\begin{tabular}{lccccc}
\toprule
Metric & $\mathcal{SM}$ & $\mathcal{OS}$ & Stateless & NOVA & \framework \\
\midrule
\rowcolor{gray!15}
\multicolumn{6}{l}{\textsc{Cost / successful topic}} \\[-1pt]
\quad $Y_{7,7,6,7}\!\ge\!1$ &
\$6.03 &
\textit{NA} &
$\mathbf{\$5.07}$ &
\$16.32 &
\$6.52 \\
\quad $Y_{7,7,6,7}\!\ge\!2$ &
\textit{NA} &
\textit{NA} &
\$35.52 &
\$130.56 &
$\mathbf{\$22.00}$ \\
\quad $Y_{6,7,6,7}\!\ge\!1$ &
\$2.58 &
\textit{NA} &
$\mathbf{\$1.48}$ &
\$5.22 &
\$5.68 \\
\quad $Y_{6,7,6,7}\!\ge\!2$ &
\$12.05 &
\textit{NA} &
\$17.76 &
\$11.87 &
$\mathbf{\$7.65}$ \\
\quad $Y_{6,7,6,7}\!\ge\!3$ &
\textit{NA} &
\textit{NA} &
\textit{NA} &
\$130.56 &
$\mathbf{\$11.73}$ \\
\bottomrule
\end{tabular}
\end{table}

%% file: content_v2/appendix/metrics.tex
\section{Evaluation Rubrics and Algorithms}
\label{app:metrics}

\input{content_v2/tables/axes}

\input{content_v2/tables/nob_axes}

\input{content_v2/tables/nob}

\input{content_v2/tables/soundness}

\input{content_v2/tables/clarity}

\input{content_v2/tables/feasibility}

\input{content_v2/tables/significance}

\input{content_v2/tables/holistic}

\begin{algorithm}[h]
\caption{Exact Novel-Sound-Clear Count: $Y_{k,l,m,\tau}$}
\label{alg:yield}
\begin{algorithmic}[1]
\Require Idea set $\mathcal{I}$;
         non-obviousness scores $\{\NB_i\}$; 
         soundness scores $\{\SO_i\}$;
         pairwise distinctness scores $\{\DIV_{ij}\}$;
         thresholds: $k$ (non-obviousness), $l$ (soundness), $m$ (clarity), $\tau$ (distinctness)
\Ensure  $\mathcal{K}^\star$, a \textit{maximum-cardinality} \textit{mutually distinct} eligible set

\State $\mathcal{N} \leftarrow \{\, i \in \mathcal{I} :  \NB_i \ge k \ \wedge\ \SO_i \ge l \, \wedge \CL_i \ge m \}$
\State $E \leftarrow \{\, (i,j) \in \mathcal{N}\times\mathcal{N} : \DIV_{ij} \ge \tau \,\}$
\State $\mathcal{K}^\star \leftarrow \emptyset$

\Function{Expand}{$\mathcal{K}, \mathcal{C}$}
    \If{$|\mathcal{K}| > |\mathcal{K}^\star|$}
        \State $\mathcal{K}^\star \leftarrow \mathcal{K}$
    \EndIf
    \If{$|\mathcal{K}| + |\mathcal{C}| \le |\mathcal{K}^\star|$}
        \State \Return
    \EndIf
    \For{$v \in \mathcal{C}$, in order}
        \If{$|\mathcal{K}| + |\mathcal{C}| \le |\mathcal{K}^\star|$}
            \State \textbf{break}
        \EndIf
        \State $\mathcal{C}' \leftarrow \{\, u \in \mathcal{C} : u \ne v,\ (v,u) \in E \,\}$
        \State \Call{Expand}{$\mathcal{K} \cup \{v\},\ \mathcal{C}'$}
        \State $\mathcal{C} \leftarrow \mathcal{C} \setminus \{v\}$
    \EndFor
\EndFunction

\State \Call{Expand}{$\emptyset,\ \mathcal{N}$}
\State \Return $\mathcal{K}^\star$
\end{algorithmic}
\end{algorithm}

%% file: content_v2/tables/axes.tex
\begin{table}[H]
\centering
\small
\caption{The eight diversity axes. The core axes are marked with $\dagger$.}
\label{tab:axes}
\begin{tabularx}{\linewidth}{@{}l X@{}}
\toprule
\textbf{Axis} & \textbf{What it asks} \\
\midrule
Failure Mode$^{\dagger}$        & What problem is solved?              \\
Causal Diagnosis$^{\dagger}$    & Why does the problem occur?          \\
Intervention$^{\dagger}$        & What is the central mechanism?       \\
Signal Source                   & What measurable signal is used?      \\
Intervention Locus              & Where in the system does it act?     \\
Objective                       & What tradeoff is optimised?          \\
Evaluation Regime               & Where would it be tested?            \\
Assumption Class                & What hidden assumption must hold?    \\
\bottomrule
\end{tabularx}
\end{table}

%% file: content_v2/tables/nob_axes.tex
\begin{table}[H]
\centering
\small
\caption{Three non-obviousness axes.}
\label{tab:nob_axes}
\begin{tabularx}{\linewidth}{@{}l X@{}}
\toprule
\textbf{Axis} & \textbf{What it asks} \\
\midrule
\textit{Problem--Mechanism Pairing} & Is the mechanism chosen the obvious/default fix for the stated problem, or a non-default one? \\
\textit{Causal Diagnosis} & Is the underlying cause of the failure being addressed a standard, expected one, or a novel diagnosis? \\
\textit{Intervention} & Is the point in the pipeline/system where the fix is applied a conventional one, or an unusual one reviewers rarely consider? \\
\bottomrule
\end{tabularx}
\end{table}

%% file: content_v2/tables/nob.tex
\begin{table}[H]
\centering
\small
\caption{Non-obviousness rubric. The accepted region used in our study is shown in \textcolor{green}{green}.}
\label{tab:nob}
\begin{tabularx}{\linewidth}{@{}c X@{}}
\toprule
$\NB$ & Meaning \\
\midrule
0 & \textit{Obvious}: a standard fix; the first thing a reviewer would propose. \\
1 & Near-trivial variant of a known fix; no new reasoning. \\
2 & \textit{Straightforward extension}: familiar components combined in the expected way. \\
3 & Predictable combination of two known ideas. \\
4 & \textit{Mildly non-obvious}: an interesting framing, but the core move is still the default. \\
5 & {\textit{Moderately non-obvious}: a real failure mode paired with a non-default mechanism, still reachable by careful reasoning.} \\
6 & \textcolor{green}{\textit{Solidly non-obvious}}: a meaningful failure mode and a mechanism that is not the obvious one. \\
7 & \textcolor{green}{\textit{Notably non-obvious}}: a surprising problem--mechanism pairing or an unusual signal source. \\
8 & \textcolor{green}{\textit{Highly non-obvious}}: a novel causal diagnosis, or a mechanism acting where reviewers rarely look. \\
9 & \textcolor{green}{\textit{Exceptional}}: a reframing of the problem itself. \\
\bottomrule
\end{tabularx}
\end{table}

%% file: content_v2/tables/soundness.tex
\begin{table}[H]
\centering
\small
\caption{Soundness rubric. The \textit{severe} cases are shown in \textcolor{red}{red}, and the accepted region is in \textcolor{green}{green}.}
\label{tab:soundness}
\begin{tabularx}{\linewidth}{@{}c X@{}}
\toprule
$\SO$ & Meaning \\
\midrule
0 & \textcolor{red}{\textit{Fatally flawed}}: the mechanism contradicts itself or relies on something unavailable in its own setting. \\
1 & \textcolor{red}{\textit{Mostly unsound}}: the core causal claim doesn't follow from its own premises without unstated extra leaps. \\
2 & \textcolor{red}{\textit{Significant gap}}: the mechanism is plausible in isolation but rests on an assumption very likely false or untested. \\
3 & \textcolor{red}{\textit{Shaky}}: reasoning holds in the common case but ignores an obvious failure mode of its own mechanism. \\
4 & \textit{Mostly sound with a caveat}: the core logic holds, but at least one non-trivial edge case or confound is unaddressed. \\
5 & \textit{Sound with minor gaps}: the mechanism is logically coherent; a few secondary assumptions are untested but plausible. \\
6 & \textcolor{green}{\textit{Solid}}: the causal chain from mechanism to claimed effect is clear and internally consistent, with reasonable, explicitly stated assumptions. \\
7 & \textcolor{green}{\textit{Rigorous}}: the reasoning anticipates and addresses its own likely failure modes or confounds. \\
8 & \textcolor{green}{\textit{Highly rigorous}}: plausible alternative explanations for the claimed effect are explicitly considered and ruled out by the mechanism's own design. \\
9 & \textcolor{green}{\textit{Airtight}}: the mechanism follows from well-justified premises with no exploitable logical or technical gap; a skeptical reviewer finds no wedge to attack the core reasoning. \\
\bottomrule
\end{tabularx}
\end{table}

%% file: content_v2/tables/clarity.tex
\begin{table}[H]
\centering
\small
\caption{Mechanism clarity rubric. The accepted region is shown in \textcolor{green}{green}.}
\label{tab:clarity}
\begin{tabularx}{\linewidth}{@{}c X@{}}
\toprule
$\CL$ & Meaning \\
\midrule
0 & \textit{Empty / hand-wavy}: no actual mechanism given, just a goal restated as a method. \\
1 & \textit{Vague direction}: names a technique family without saying what changes or how. \\
2 & \textit{Underspecified}: identifies a component to modify but not what signal, rule, or update it uses. \\
3 & \textit{Sketch-level}: the mechanism's general shape is stated but key steps are asserted rather than defined. \\
4 & \textit{Partially specified}: most of the mechanism is concrete; one central step remains hand-wavy or unresolved. \\
5 & \textit{Adequately specified}: the mechanism's main steps are all named and connected, though exact operationalization (thresholds, formulas) is left open. \\
6 & \textcolor{green}{\textit{Well specified}}: each step is concrete enough that an implementer would only need to choose minor free parameters. \\
7 & \textcolor{green}{\textit{Precisely specified}}: inputs, the transformation applied, and outputs are all stated; no structural ambiguity remains. \\
8 & \textcolor{green}{\textit{Fully operational}}: could be implemented directly from the description; only routine engineering choices are left. \\
9 & \textcolor{green}{\textit{Unambiguous and complete}}: every step, signal, and decision point is explicit enough that two independent implementers would build functionally the same thing. \\
\bottomrule
\end{tabularx}
\end{table}

%% file: content_v2/tables/feasibility.tex
\begin{table}[H]
\centering
\small
\caption{Feasibility rubric.}
\label{tab:feasibility}
\begin{tabularx}{\linewidth}{@{}c X@{}}
\toprule
$\phi$ & Meaning \\
\midrule
0 & \textit{Practically impossible}: requires resources, data, or capabilities that don't exist and aren't obtainable. \\
1 & \textit{Extremely demanding}: would require large, dedicated infrastructure or data collection beyond a typical project's scope. \\
2 & \textit{Very demanding}: feasible only with substantial new tooling, data collection, or compute beyond standard academic resources. \\
3 & \textit{Demanding}: needs meaningful new infrastructure (e.g. a new training pipeline or annotated dataset) not implied by the idea itself. \\
4 & \textit{Moderately demanding}: buildable with standard resources but requires nontrivial engineering effort or a moderately large compute budget. \\
5 & \textit{Feasible with effort}: implementable with existing tools/data with a reasonable, if nontrivial, engineering effort. \\
6 & \textit{Fairly feasible}: mostly reuses existing infrastructure/data; one or two components need custom work. \\
7 & \textit{Feasible}: could be built and tested by a small team using standard, available tools and data with modest effort. \\
8 & \textit{Highly feasible}: implementable quickly by adapting existing pipelines with minor modification. \\
9 & \textit{Immediately actionable}: testable essentially off-the-shelf, with data/tools/compute already at hand. \\
\bottomrule
\end{tabularx}
\end{table}

%% file: content_v2/tables/significance.tex
\begin{table}[H]
\centering
\small
\caption{Significance rubric.}
\label{tab:significance}
\begin{tabularx}{\linewidth}{@{}c X@{}}
\toprule
$\zeta$ & Meaning \\
\midrule
0 & \textit{Negligible}: even complete success would not meaningfully change anything a practitioner or researcher cares about. \\
1 & \textit{Marginal}: a small, local improvement noticeable only in narrow or synthetic settings. \\
2 & \textit{Minor}: a modest improvement on a narrow problem with little downstream consequence. \\
3 & \textit{Limited}: addresses a real but narrow problem; success would matter only to a small niche. \\
4 & \textit{Moderate}: meaningfully improves a recognized problem, though the problem itself is not central to the field. \\
5 & \textit{Notable}: success would meaningfully move a problem that a nontrivial part of the field actively cares about. \\
6 & \textit{Significant}: success would change how a meaningful subset of practitioners approach a known, important problem. \\
7 & \textit{Important}: success would resolve or substantially advance a problem widely recognized as a major open issue. \\
8 & \textit{Major}: success would shift how the field thinks about the underlying failure mode itself, beyond just improving numbers. \\
9 & \textit{Field-defining}: success would open or redefine a research direction, changing what subsequent work in the area targets. \\
\bottomrule
\end{tabularx}
\end{table}

%% file: content_v2/tables/holistic.tex
\begin{table}[H]
\centering
\small
\caption{Holistic pairwise distinctness rubric ($\DIV_{ij}$). The accepted region is shown in \textcolor{green}{green}.}
\label{tab:holistic}
\begin{tabularx}{\linewidth}{@{}c X@{}}
\toprule
$\DIV$ & Meaning \\
\midrule
0 & \textit{Semantic duplicate}: same research claim and substantive mechanism. \\
1 & \textit{Cosmetic rewrite}: only wording, presentation, or naming differs. \\
2 & \textit{Minor variant}: same core mechanism and causal story with a small operational change. \\
3 & \textit{Meaningful variant}: meaningful modification, but still the same research direction. \\
4 & \textit{Sibling direction}: one core component changes; recognizable common direction remains. \\
5 & \textit{Partially distinct}: substantial change, but mechanism-family/core overlap remains. \\
6 & \textit{Distinct direction}: separately actionable direction with a genuinely different central mechanism or causal route; contextual differences alone never qualify. \\
7 & \textcolor{green}{\textit{Clearly distinct}}: mechanism, diagnosis, and evidence path are clearly separated. \\
8 & \textcolor{green}{\textit{Highly distinct}}: only the broad problem area or domain substantially connects the ideas. \\
9 & \textcolor{green}{\textit{Orthogonal}}: essentially all substantive components differ; reserve this for rare cases. \\
\bottomrule
\end{tabularx}
\end{table}

%% file: content_v2/appendix/dataset.tex
\FloatBarrier
\section{Background Corpus Collection}
\label{app:corpus-collection}

\begin{algorithm}[H]
\caption{Greedy Max-Min Topic Diversity Selection}
\label{alg:greedy_diversity}
\begin{algorithmic}[1]
\Require Topic names $\{t_1, \ldots, t_N\}$, sentence encoder $\mathcal{F_{\theta}}$, budget $\Gamma$
\Ensure Selected set $\mathcal{S}$, $|\mathcal{S}| = \Gamma$
\State $\hat{e_i} \gets \mathcal{F_{\theta}}(t_i)$ $\forall i$ \Comment{encode topic names (normalized vectors)}
\State $S \gets \hat{E}\hat{E}^\top$ where $\hat{E} = [\hat{e}_1, \ldots, \hat{e}_N]^\top$ \Comment{cosine similarity matrix}
\State $s^* \gets \arg\min_{i} \frac{1}{N-1} \sum_{j \neq i} S_{ij}$ \Comment{most isolated topic as seed}
\State $\mathcal{S} \gets \{s^*\}$, \quad $\mathcal{R} \gets \{1,\ldots,N\} \setminus \{s^*\}$ \Comment{initialise selected and remaining sets}
\While{$|\mathcal{S}| < \Gamma$}
    \State $i^* \gets \arg\min_{i \in \mathcal{R}} \max_{j \in \mathcal{S}}\, S_{ij}$ \Comment{pick topic least similar to any selected}
    \State $\mathcal{S} \gets \mathcal{S} \cup \{i^*\}$, \quad $\mathcal{R} \gets \mathcal{R} \setminus \{i^*\}$ \Comment{update selected and remaining sets}
\EndWhile
\State \Return $\mathcal{S}$ 
\end{algorithmic}
\end{algorithm}

\input{content_v2/tables/domains_topics}

%% file: content_v2/tables/domains_topics.tex
\begin{table}[H]
\centering
\small
\caption{List of domains and topics included in our dataset.}
\label{tab:domains_topics}
\begin{tabular}{@{}ccl@{}}
\toprule
\multicolumn{1}{l}{\textbf{Domain}} & \multicolumn{1}{l}{\textbf{\# BKDG}} & \multicolumn{1}{c}{\textbf{Topic Name}}                  \\ \midrule
\multirow{4}{*}{\textit{cs.AI}}     & 8                                    & Scaling LLM Reasoning via Reinforcement Learning         \\
                                    & 8                                    & Scaling LLM Reasoning via Reinforcement Learning         \\
                                    & 5                                    & Reasoning in Large Language Models                       \\
                                    & 6                                    & Reinforcement Learning for Advanced AI Reasoning         \\ \midrule
\multirow{6}{*}{\textit{cs.CL}}     & 8                                    & Deep Learning Representation and Analysis                \\
                                    & 8                                    & Scaling Laws for Neural Networks                         \\
                                    & 7                                    & Knowledge Localization and Model Editing                 \\
                                    & 8                                    & Efficient Attention Mechanisms and KV Cache Optimization \\
                                    & 6                                    & Multilingual Multitask Language Understanding Evaluation \\
                                    & 6                                    & Adversarial Robustness of LLM Safety Alignment           \\ \midrule
\multirow{6}{*}{\textit{cs.CV}}     & 8                                    & Efficient Long-Range Dependency Modeling                 \\
                                    & 8                                    & Embodied AI and Autonomous Systems                       \\
                                    & 7                                    & Controllable Text-to-Image Synthesis and Editing         \\
                                    & 8                                    & Adaptive and Personalized Generative Modeling            \\
                                    & 6                                    & Deep Learning for Image Anomaly Detection                \\
                                    & 6                                    & Multi-camera Bird's-Eye-View 3D Object Detection         \\ \midrule
\multirow{2}{*}{\textit{cs.DC}}     & 7                                    & Efficient Large Language Model Serving Systems           \\
                                    & 7                                    & Scalable LLM Training and Inference Systems              \\ \midrule
\textit{cs.IR}                      & 7                                    & Autoregressive Generative Modeling Across Modalities     \\ \midrule
\multirow{6}{*}{\textit{cs.LG}}     & 8                                    & Large-Scale Model Parameter Optimization                 \\
                                    & 7                                    & Offline Reinforcement Learning and Generative Modeling   \\
                                    & 7                                    & Geometry and Linearity of Weight Space                   \\
                                    & 6                                    & Unifying Diffusion and Flow-Based Generative Models      \\
                                    & 6                                    & Efficient LLM Compression and Adaptation                 \\
                                    & 5                                    & Foundation Models for Time Series Forecasting            \\ \midrule
\multirow{6}{*}{\textit{cs.RO}}     & 8                                    & Diffusion Models for End-to-End Autonomous Driving       \\
                                    & 8                                    & Robot Manipulation Policy Learning                       \\
                                    & 7                                    & End-to-End Autonomous Driving                            \\
                                    & 7                                    & Imitation Learning for Robotic Manipulation              \\
                                    & 7                                    & Visuomotor Learning for Robot Manipulation               \\
                                    & 5                                    & Diffusion-based Generative Modeling                      \\ \midrule
\textit{cs.SE}                      & 7                                    & RL and Search for LLM Reasoning                          \\ \bottomrule
\end{tabular}
\end{table}

%% file: content_v2/appendix/alpharesearch.tex
\FloatBarrier
\section{Extension to Verifiable Algorithm Discovery}
\label{app:alpharesearch}

\paragraph{Adapting search to \alpharesearch and \alebench}
\begin{enumerate}
    \item Since there are no background papers for each task, we replace them with the task definition and runtime constraints. Each task also has with a trivial/standard solution implemented by \opusnew, which serves as a baseline for the quality of the generated ideas, and all the upcoming ideas are expected to beat it. 
    \item Each idea generated by the Ideator now consists of a proposal and the concrete Python/C++ program that implements the idea. The program is then run for 60s during the search and 300s in the final evaluation on \alpharesearch, and within the contest's time limit (2--3s) on each of the 50 public test cases on \alebench, to generate its output, and evaluate the quality according to the task-specific metric. In case the program has any syntax or coding errors, we have a separate loop with the Ideator to allow at most 10 superficial fixes, which are independent of the QD search budget. To ensure the fixes do not alter the logic of the program, we have precise instructions for the Ideator and limit the change to a maximum of 20 lines. A generated program is \textit{valid} if it runs and generates an output in the allotted time on every instance, the output follows the required format, and passes the benchmark's checker, along with our stricter checks on \alpharesearch. An \textit{invalid} program is assigned $Q_i=0$. 
    \item As each task may have a separate task quality metric on its own scale, they may not be directly comparable. So, we normalize the score of each idea by the baseline score of the trivial solution and a reference score provided in the benchmark, i.e., the \textit{human-best} on \alpharesearch and the mean score of the \textit{contest-winner} on \alebench. Formally, for an instance $j$ of a task $t$, let $s^0_j$ be the score of the trivial solution, and $s^{\star}_j$ be the reference score. For an idea $i$ with score $s_{ij}$, we define the normalized progress percentage as $Q_i = \frac{100}{|J_t|} \sum_{j \in J_t} \max\Big(0, \frac{s_{ij} - s^0_j}{s^{\star}_j - s^0_j}\Big)$, where $J_t$ is the set of instances of $t$ (one or two on \alpharesearch, and 50 test cases on \alebench). Under this formulation, an idea that does not beat the trivial solution will have a progress of 0, an idea that equals or surpasses the best score provided will be marked $\geq 100$. 
    \item We remove the clarity and non-obviousness judges, as the Ideator directly generates a formal program, and the progress metric is sufficient to estimate the difference with respect to a trivial solution. Further, the soundness panel is reduced to a single binary \textit{fatal} decision which checks if the program is aligned with the proposal, and does not contain any shortcuts, hard-coded solutions, incorrect assumptions, or logic. The diversity-judge and archives remain as is, but since the \textit{problem statement} and \textit{expected effect} are fixed, the diversity is mainly associated with the \textit{central mechanism}. The critic also remains and provides feedback to the Ideator based on the comments from the soundness judge and performance score.
    \item A fresh and \textit{valid} idea which fails to clear the baseline is allowed to be repaired once, as per the feedback from the critic, and if it still fails again, it is discarded. An idea that clears the baseline is allowed at most two refinements, and as usual, a higher quality child replaces its parent, or closest incumbent.
    \item Since quality is now a verified score, the final evaluation to compute the \yield employs an LLM-\textit{jury} to assess if the central mechanisms provided in two ideas are truly different, i.e., a pair of ideas is assigned as distinct if a majority in the jury identifies it as such. Unlike a Likert-scale-based diversity score, which requires distinction along multiple axes for free-form research ideas, the diversity check is limited to a binary decision here, as we directly compare the procedure followed in the code.
    \item Here, due to budget and computation constraints, we resort to open-source models such as \qwen for the Ideator and \qwenmoe for the internal critic, soundness, and diversity judges. The external diversity jury includes \gemma, \minimax, and \glm, as they demonstrate strong code and reasoning abilities. 
\end{enumerate}